\pdfoutput=1
\documentclass[10pt]{article}
\usepackage{biabench-report}
\hypersetup{pdftitle={BIABench: Evaluating AI agents on real-world bioimage analysis tasks},
  pdfauthor={Zixuan Pan, Davide Panzeri, Lukas Johanns, Marilin Moor, Yu Zhou, Hedi Peterson, Yiyu Shi, Jianxu Chen}}

\graphicspath{{figures/}{./}}

\newcommand{\bench}{\textbf{BIABench}\xspace}

\reportmark{{\color{accent}BIA}{\color{markgray}Bench}}
\reportshorttitle{Evaluating AI agents on real-world bioimage analysis tasks}
\reporttitle{\bench: Evaluating AI agents on real-world bioimage analysis tasks}

\reportauthors{%
\textbf{Zixuan Pan}$^{1,*}$,
\textbf{Davide Panzeri}$^{2,*}$,
\textbf{Lukas Johanns}$^{2}$,
\textbf{Marilin Moor}$^{3}$,
\textbf{Yu Zhou}$^{2}$,
\textbf{Hedi Peterson}$^{3}$,
\textbf{Yiyu Shi}$^{1,\dagger}$,
\textbf{Jianxu Chen}$^{2,\dagger}$}

\reportaffils{%
$^{1}$Department of Computer Science and Engineering, University of Notre Dame, Notre Dame, USA\\
$^{2}$Leibniz-Institut f\"ur Analytische Wissenschaften -- ISAS -- e.V., Dortmund, Germany\\
$^{3}$Institute of Computer Science, University of Tartu, Tartu, Estonia\par\vspace{3pt}
{\small$^{*}$These authors contributed equally.\quad
$^{\dagger}$Corresponding authors: \href{mailto:yshi4@nd.edu}{yshi4@nd.edu}, \href{mailto:jianxu.chen@isas.de}{jianxu.chen@isas.de}\par}}

\reportabstract{%
Artificial-intelligence (AI) agents hold promise for automating bioimage analysis, yet no
benchmark evaluates whether they can carry out real-world analyses end to end.
Such analyses are hard for agents because 2D images, 3D volumes and time-lapse
sequences are often too large to read as context, so an agent must choose and
run an analysis through code, specialized software and rendered views.
Published studies make this capability testable, because each pairs raw images
with a peer-reviewed result. We introduce \bench, a benchmark of $16$ tasks
reconstructed from published biological studies that retain their scientific
questions, imaging data and ground truth. The tasks span eleven analysis
subtasks and modalities from H\&E histology to single-molecule localization
microscopy. Each submission receives an outcome score, which compares the output
files with the ground truth using field-standard metrics, and a process score,
in which a vision--language model judges method choice and quality control
against an expert-written rubric. We evaluated general-purpose and biology-specific agents across several language models, with repeated runs of every task. Routine two-dimensional tasks were solved well, but on some tasks that added a third dimension or a time axis no agent scored above $0.19$. Neither biological specialization,
stronger models nor detailed expert instructions closed this gap. The agents were also unreliable, with scores varying more
between repeated runs of one agent than between different agents, and without
ground truth a correct run could not be told from a wrong one by its process
score or by the time spent.
Released openly with its data and code, \bench provides a verifiable framework
for evaluating, and eventually training, agents for reliable long-horizon
bioimage analysis.}

\reportlinks{%
\linkrow{\faGlobe}{Website}{https://biabench.github.io}{https://biabench.github.io}
\linkrow{\faGithub}{Code}{https://github.com/BIABench/BIABench}{https://github.com/BIABench/BIABench}
\linkrow{\raisebox{-0.2em}{\includegraphics[height=1.1em]{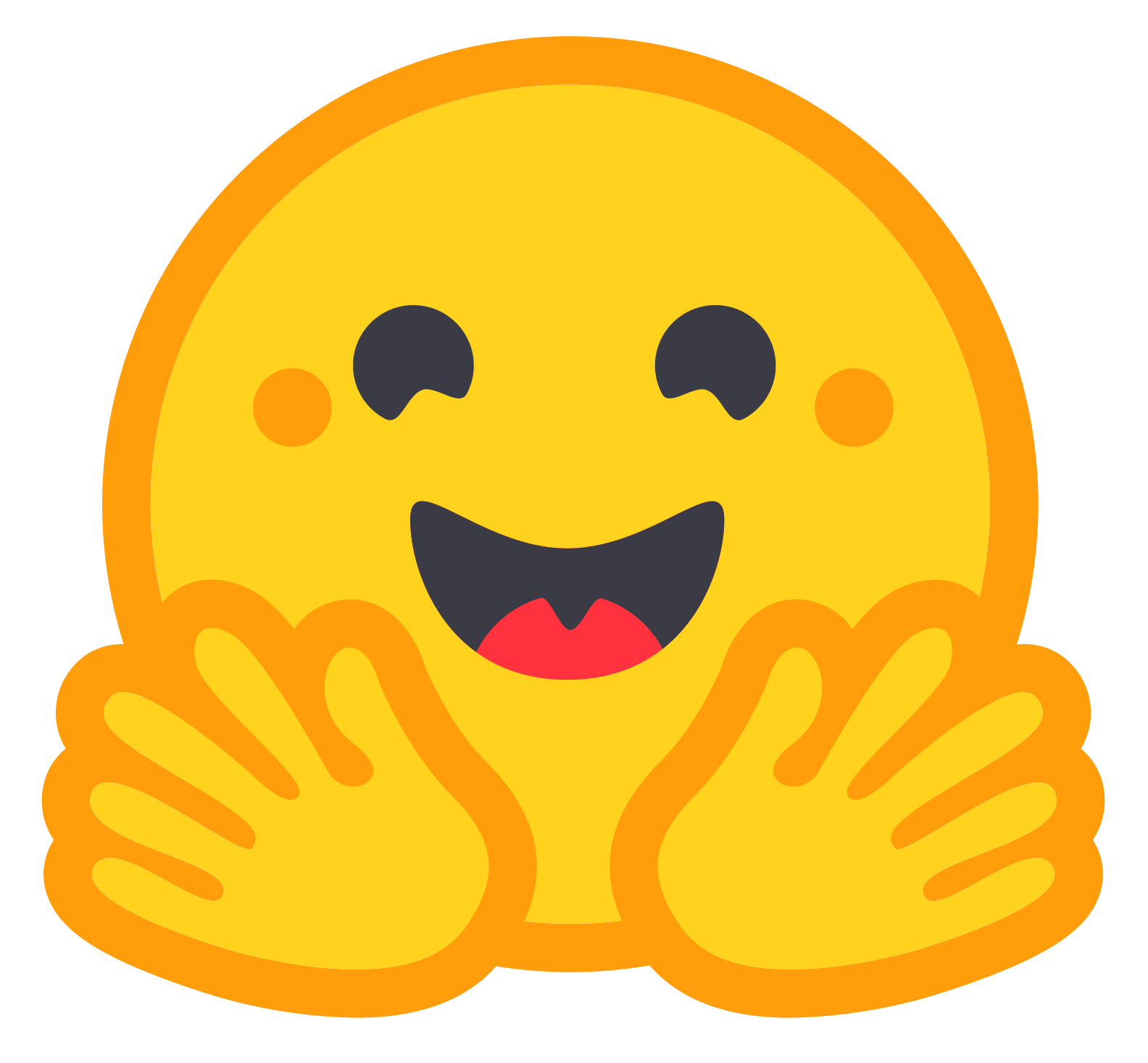}}}{Data}{https://huggingface.co/datasets/BIABench/BIABench}{https://huggingface.co/datasets/BIABench/BIABench}}

\reportlogo{figures/biabench_logo.png}

\begin{document}

\makereporttitle

\section{Introduction}\label{sec:intro}

Bioimage analysis is central to how biologists turn imaging data into quantitative measurements and biological insight. However, even routine analyses, from counting labeled cells to tracking them over time, require specialist software and programming expertise that is scarce in many biology laboratories~\citep{omega,bioimageiochatbot}. Recent advances in large language models (LLMs) have made artificial intelligence (AI) agents increasingly capable of planning and executing complex tasks, raising the
possibility of automating such analyses~\citep{MLLMChen}. In principle, an agent could take images and an instruction, plan the analysis, execute the necessary tools and report the result. Early examples are beginning to realize this vision by enabling agents to interact with established bioimage-analysis software~\citep{omega,agenticj,copilotj}. \emph{Yet whether current agents can reliably perform real-world, end-to-end bioimage analyses from raw images to biological insights remains unclear, as no benchmark systematically evaluates this capability.}

Benchmarks for bioimage analysis have so far focused on evaluating models on a specific step or a specific task of an analysis, such as nucleus segmentation~\citep{caicedo2019dsb}, cell tracking~\citep{celltrackingchallenge}, in-silico labeling~\citep{kauffmann2026insilico} or image quality control~\citep{autoqc}, and are well suited to comparing the accuracy and generalization of specific
models. In contrast, benchmarks for agents evaluate long-horizon tasks in question answering, coding and computer use~\citep{gaia,swebench,osworld}, with recent ones extending to scientific and biological
analysis~\citep{labbench,bixbench,bixbench3,biomnibench}. Those that include biological images either only test visual reasoning by multimodal models~\citep{microvqa} or ask an agent to train a model on an existing
benchmark dataset and score on its held-out predictions~\citep{rexmle,bioxarena}. Both inherit the limitation of the single-step framing, whereas a real analysis rarely stops at one step. The images may first need denoising or stitching, the question may call for colocalization, tracking or detection, and even a segmentation is only the start of the feature extraction and statistics that yield the biological result. No benchmark evaluates whether an agent can carry out this entire cycle, from raw images to the study's conclusion, and doing so poses two distinct challenges for current agents. First, bioimage data are often too
large or structurally complex to be supplied directly to an agent as context, spanning high-resolution 2D images, 3D volumes and time-lapse sequences, even multichannel 3D time-lapse. An agent must instead inspect and manipulate the data through computation, specialized software and rendered views. Second, the task is end-to-end with open-ended solutions rather than a predefined prediction problem, requiring an agent to determine
an analysis strategy from the biological question and data, execute it and produce the required scientific outputs. An agent may call established tools such as Cellpose~\citep{cellpose} or StarDist~\citep{stardist}, write custom code, or operate graphical user interface (GUI) software, such as Fiji~\citep{fiji} or napari~\citep{napari}.

Here we introduce \bench, a benchmark for evaluating autonomous AI agents on real-world, end-to-end bioimage analysis (Fig.~\ref{fig:design}). Its $16$ tasks are reconstructed from published biological studies, retaining their scientific questions, imaging data and reported outputs as ground truth. The agent receives the raw images and the biologist's instruction, and its output is scored against the study's original peer-reviewed result. The tasks cover eleven analysis subtasks, span high-resolution 2D images, 3D volumes and time-lapse sequences, and include modalities ranging from H\&E histology to
single-molecule localization microscopy (Fig.~\ref{fig:design}b). The samples span six source organisms, from bacteria to humans, the imaged
structures range from single molecules to cell monolayers, and the input data total
$13.9$\,GB (Fig.~\ref{fig:design}c). Because an analysis can run to completion and produce plausible figures yet reach the wrong result, each submission receives two scores (Fig.~\ref{fig:design}a). First, an outcome score compares required outputs, such as masks, tracks and tables, with the study's ground truth using field-standard metrics. Second, a process score evaluates method choice, quality control, figures and documentation against expert-written rubrics using a vision--language model (VLM), with quality verified by human experts (Section~\ref{sec:judge}).

\newcommand{\panelletter}[1]{\sffamily\bfseries\fontsize{8}{9}\selectfont #1}
\begin{figure*}[p]
\centering
\newlength{\panelh}
\newlength{\figonew}\setlength{\figonew}{0.97\textwidth}
\settoheight{\panelh}{\includegraphics[width=\figonew]{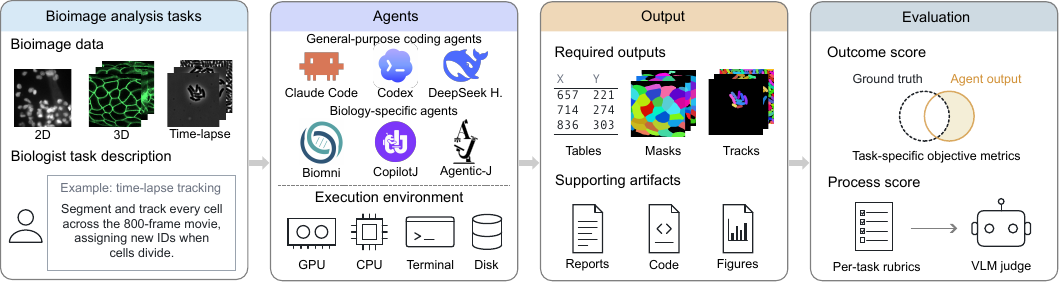}}%
\makebox[\textwidth][c]{\llap{\raisebox{\dimexpr\panelh-1ex\relax}{\panelletter{a}}\hspace{4pt}}\includegraphics[width=\figonew]{overview-v4.pdf}}\par
\vspace{2pt}
\settoheight{\panelh}{\includegraphics[width=\figonew]{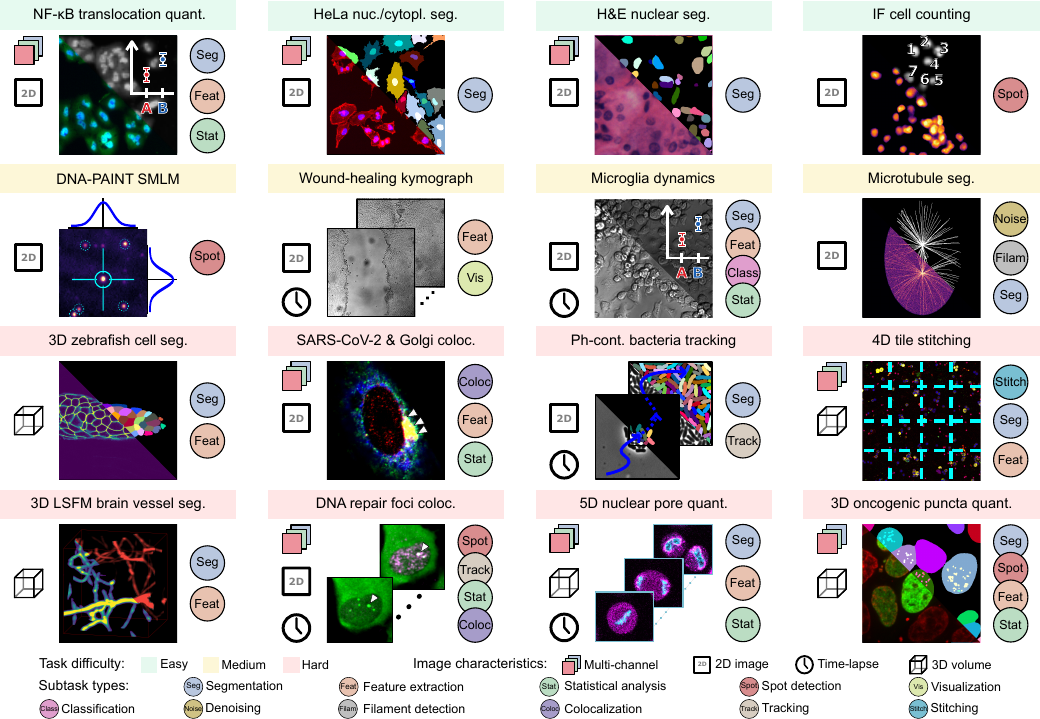}}%
\makebox[\textwidth][c]{\llap{\raisebox{\dimexpr\panelh-1ex\relax}{\panelletter{b}}\hspace{4pt}}\includegraphics[width=\figonew]{bioimage-bench-v12.pdf}}\par
\vspace{2pt}
\settoheight{\panelh}{\includegraphics[width=\figonew]{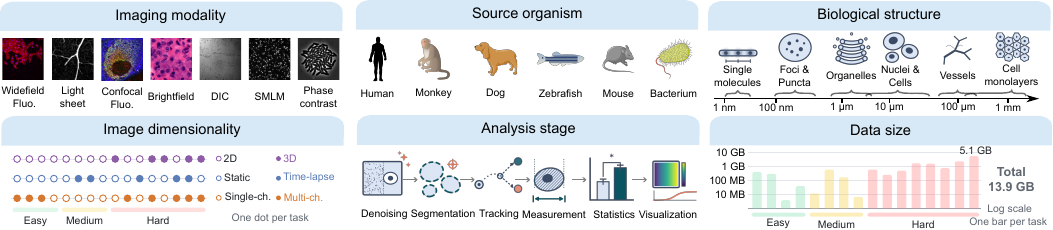}}%
\makebox[\textwidth][c]{\llap{\raisebox{\dimexpr\panelh-1ex\relax}{\panelletter{c}}\hspace{4pt}}\includegraphics[width=\figonew]{data_stat-v3.pdf}}\par
\vspace{-3pt}%
\caption{\textbf{Design of \bench.}
\textbf{a)} Benchmark workflow. Each task pairs a published study's raw images with a
biologist's instruction (i.e., prompt) and asks for the study's own readout, with the ground truth
withheld from the agent. Agents run in a shared, controlled environment. An \textit{outcome score}
compares the required output files (e.g., tables, masks, tracks) with the ground truth using
task-specific metrics (zero for missing files), and a \textit{process score} uses a
vision--language model to rate supporting artifacts (e.g., reports, code, figures) against a
severity-weighted checklist of the task's rubric items.
\textbf{b)} The $16$ tasks, each rebuilt from a published study, showing a
representative input with its ground truth or output, analysis
subtasks (circles) and image properties (badges). Appendix~\ref{app:tasks} gives provenance, and Tables~\ref{tab:tasks} and~\ref{tab:taskspec} list subtasks, outputs and metrics.
\textbf{c)} Diversity of the task suite by imaging modality, source organism,
biological structure (ordered by approximate size), image dimensionality,
analysis stage and input data size (log scale). ch., channel; Fluo., fluorescence.}
\label{fig:design}
\end{figure*}

Our contributions are as follows.
\begin{itemize}
\item \textbf{A recipe for turning published studies into verifiable tasks.}
  Each task reduces a published study to its raw images, the biologist's
  request for the study's key readout and the reported outputs as ground truth.
  The same recipe can extend the benchmark to new studies.
\item \textbf{A benchmark of end-to-end bioimage analysis.} $16$ tasks spanning
  eleven analysis subtasks and 2D, 3D and time-lapse data, each specified by a
  brief and a detailed instruction and scored on both its outcome and its
  process. The agent chooses the method and tools, including GUI software, and
  an agent-agnostic interface provides wrappers for six agents of three kinds (an
  LLM tool-use agent, coding command-line agents and GUI agents that operate
  Fiji/ImageJ).
\item \textbf{A comprehensive empirical study of current agents.} We evaluate both
  general-purpose and biology-specific agents across several language models and
  instruction levels, and locate the remaining gap to expert-level bioimage analysis in data complexity, reliability
  and self-checking.
\end{itemize}

\section{Related work}\label{sec:related}

\paragraph{General-purpose agent benchmarks.}
LLM agents are tracked by benchmarks that run the agent in an
environment and score the state it leaves behind. SWE-bench runs a repository's own test
suite against an agent's patch~\citep{swebench}, GAIA poses assistant tasks with short,
verifiable answers~\citep{gaia}, and AgentBench probes tool use across many
environments~\citep{agentbench}. More recent suites extend this execution-grounded
principle to new action spaces: $\tau$-bench and its successor $\tau^2$-bench score
policy-constrained tool--agent--user interaction~\citep{taubench,tau2bench}, OSWorld
evaluates computer-use agents by their effect on a real operating
system~\citep{osworld}, Terminal-Bench measures long-horizon work in containerized
terminals~\citep{terminalbench}, BrowseComp measures persistent web browsing~\citep{browsecomp},
MLE-bench grades machine-learning engineering against human Kaggle
leaderboards~\citep{mlebench}, and StartupBench scores the finished work
products of professional workflows drawn from commercially adopted AI services~\citep{startupbench}. From this line of work we take two
design choices. First, each task is defined by the files it must produce. Second, as the \emph{harness}
can move success rates as much as the model
does~\citep{clawswebench,harnessbench}, we evaluate every agent through one shared
interface, recover output files from the filesystem, and treat each vendor
coding command-line tool as a model-confounded harness. We differ in the target and the task design. Every task is
a real-world bioimage-analysis problem taken from a published biological study and posed as
a biologist would pose it, and we score the objective quantity the study itself reports
(Dice, Jaccard, the Cell Tracking Challenge TRA measure, the Kolmogorov--Smirnov
statistic, and Pearson correlation) on real microscopy.

\paragraph{AI agents for biology.}
A growing number of agents act as autonomous biological collaborators. For
hypothesis generation and design, Google's AI co-scientist proposes and refines biomedical
hypotheses through a multi-agent generate--debate--evolve process~\citep{aicoscientist},
and the Virtual Lab orchestrates a team of language-model ``scientists'' that designed
experimentally validated SARS-CoV-2 nanobodies~\citep{virtuallab}. For the analysis loop,
general biomedical agents retrieve tools, write code and run analyses. Biomni's
action-discovery agent mines tools, databases and protocols from the literature
across $25$ domains, its harness adds $6$ to $12$ points over the bare model on
text-based biomedical questions, it has been exercised in wet-lab case studies,
and its authors list image inputs as a future direction~\citep{biomni}; the
self-evolving STELLA does the same~\citep{stella}. A first wave of agents
now targets bioimage analysis directly, across every major interface: in napari, Omega
holds a conversation while it segments and quantifies~\citep{omega}; for ImageJ/Fiji,
Agentic-J~\citep{agenticj} and CopilotJ~\citep{copilotj} drive a live session from natural
language; GenCellAgent routes between Cellpose, micro-SAM, and other tools for cellular
segmentation~\citep{gencellagent}; the BioImage.IO Chatbot connects users to the model zoo
and runs its tools~\citep{bioimageiochatbot}; and LSM-Copilot packages microscopy skills
that run unchanged across several host agents~\citep{lsmcopilot}. \bench draws its biology-specific agents-under-test from this landscape and adds general-purpose coding agents, spanning the three interface classes: the Python tool-use agent Biomni, three coding command-line agents (Claude Code, Codex and DeepSeek Harness), and two GUI agents that operate Fiji/ImageJ (CopilotJ and Agentic-J).

\paragraph{Task-specific bioimage benchmarks.}
The bioimage community has long benchmarked models on single analysis steps
against ground truth. The 2018 Data Science Bowl scored nucleus segmentation
across imaging experiments~\citep{caicedo2019dsb}, the Cell Tracking Challenge
has ranked segmentation and tracking methods for a
decade~\citep{celltrackingchallenge}, recent collections benchmark in-silico
labeling, the prediction of fluorescence from transmitted-light
images~\citep{kauffmann2026insilico}, and AutoQC-Bench benchmarks quality
control of high-throughput microscopy~\citep{autoqc}. These resources evaluate specific models on one step, and several \bench tasks draw on datasets of this kind. Because a biologist's analysis continues past that step (Section~\ref{sec:intro}), \bench scores an agent on the whole cycle, with the outcome
taken at the study's endpoint and the process rubric covering every stage from
loading the raw data to reporting.

\paragraph{Benchmarks in biology.}
Existing biomedical agent benchmarks score quantities other than an
end-to-end physical measurement. A first group scores text answers, multiple-choice
reasoning, or generated code: LAB-Bench and its successor grade biology-research
questions~\citep{labbench,labbench2}, BixBench poses open-ended computational-biology
analyses judged against ground-truth answers~\citep{bixbench}, and, for microscopy specifically, MicroVQA probes
expert visual reasoning and hypothesis generation~\citep{microvqa}. A newer line grades the output files of
end-to-end bioinformatics pipelines against expert
references~\citep{bioagentbench,promptbiobench}, still without microscopy
tasks.

BixBench3 carries this line to the scale of whole studies. Published omics
studies are decomposed into intermediate data artifacts, an agent given the
research objective, the study's own method guidance and the raw data must
regenerate them, and each artifact is graded programmatically (identifier F1
and Lin's concordance) against the published one, with a pass threshold
calibrated on expert ratings~\citep{bixbench3}. Frontier models under one
harness reproduced fewer than half of the artifacts. Scores fell as the
analysis chain lengthened and the data grew, the best models were among the
cheapest, and the lowest-scoring attempts were marked by premature termination,
retry loops and placeholder outputs. \bench shares its principle of grading the delivered
artifact against the study's own result, and several of its observations
reappear here on microscopy in a different form, namely outcome unrelated to the
effort spent, collapse with dimensionality, and runs that end with a claim of
completion. It differs in what is left to the agent. BixBench3 prescribes the
method so that its artifacts can be matched, and its authors note that it
therefore does not test which analysis to run; our brief instruction is the
biologist's description, and the choice of tool, including GUI software, is
part of what is scored. It also compares models under a single harness with
one run per task, whereas we vary harness and model and repeat every
configuration--task pair three times.

A second group does score imaging against ground-truth metrics but frames the task as model building.
ReX-MLE~\citep{rexmle}, BioXArena~\citep{bioxarena} and
BioML-bench~\citep{biomlbench} have agents train models on existing medical and
biomedical imaging datasets and score their held-out predictions. Underlying both, the bioimage community maintains the mature,
ground-truth-scored methods an agent is expected to orchestrate, including
Cellpose~\citep{cellpose} and StarDist~\citep{stardist} for segmentation and
Fiji~\citep{fiji} with TrackMate~\citep{trackmate} for tracking; each is
evaluated as a single algorithm on a curated dataset, as in the task-specific
benchmarks above.

The concurrent BiomniBench grades the full analytical trajectory of biomedical data-analysis agents
against expert-authored ordinal rubrics applied by a validated LLM judge, and reports
that the agent harness moves scores by more than a model generation~\citep{biomnibench}.
It is a process-level complement to our design. Its score is entirely judge-derived,
whereas \bench ranks on deterministic metrics computed against ground truth and reports
the judged process score separately, with its human agreement measured; and its tasks are
tabular omics analyses under coding harnesses, whereas ours are executed microscopy
measurements that additionally exercise GUI and domain-specialized agents.
BiomniBench motivates process-level scoring with two failures of final-answer
matching: a correct answer can arise from memorization or chance, and valid
alternative analyses are marked wrong for differing from the reference. In
imaging the second failure is weaker, as the artifacts an agent delivers
describe the specimen rather than the implementation, and outputs from
different valid pipelines can be scored against common ground-truth masks and
tracks. The first is addressed by the separate process score and by an instruction that forbids
retrieving the source publication. Within bioimage analysis itself, LLM code
generation has been benchmarked at the function level with unit
tests~\citep{humanevalbia}. None of this prior work asks
whether an agent that must choose its own tools and run a complete pipeline reports the
right physical quantity on real microscopy. \bench scores agents against ground truth across many subtasks, modalities
and dimensionalities, with three runs per configuration and task.

\section{The \bench benchmark}\label{sec:benchmark}

\subsection{Tasks}\label{sec:tasks}
\paragraph{Sources and coverage.}
\bench comprises $16$ tasks, each curated from a published biological study and
inheriting that study's measurement goal and ground truth. To build a task we
dissected the study into its raw data, the analysis pipeline its authors
applied and the quantity on which its conclusion rests, and wrote the
instruction as the biologist's request for that quantity. All source data are
openly licensed and
are redistributed with the benchmark under their original terms. The suite
spans eleven analysis subtasks (segmentation, feature extraction,
statistical plotting, spot detection, colocalization, tracking, classification,
filament extraction, denoising, multi-tile mosaic stitching and visualization)
across 2D, 3D and time-lapse acquisitions and modalities including fluorescence widefield,
confocal, spinning-disk, light-sheet, phase contrast, differential interference contrast (DIC), single-molecule
localization microscopy (SMLM), AiryScan and H\&E histology. Each task is assigned a
difficulty level. Easy tasks are single-subtask or a standard two-dimensional pipeline. Medium tasks stay two-dimensional but call for a specialized method, such as single-molecule localization, optical flow or filament extraction. Most Hard tasks combine several subtasks, and a few are hard for a single but demanding subtask such as bacterial tracking across an $800$-frame time-lapse. Each task has a machine-readable
specification (i.e., a YAML file) which declares its modality, dimensionality, temporal mode, subtasks,
the exact output format, the scoring metric, and the source accession. The suite
is laid out as a task-by-subtask matrix in Table~\ref{tab:tasks},
the required output files and scoring metric of each task in Table~\ref{tab:taskspec}, and the per-task provenance in Appendix~\ref{app:tasks};
data-reconstruction tooling is released with the benchmark. The tasks can be solved with established tools such as
Cellpose~\citep{cellpose} and StarDist~\citep{stardist} for segmentation or
Fiji~\citep{fiji} with TrackMate~\citep{trackmate} for tracking; the choice of
tools is left to the agent.

\paragraph{Instructions.}\label{sec:instructions}
Each task is specified at two levels of instruction detail. The \emph{brief}
instruction is written in the voice of a biologist describing the experiment
and the desired measurement, with minimal computational guidance. The
\emph{detailed} instruction is a protocol written by an expert in bioimage analysis on the basis of the source study's methods; it additionally sketches a recommended pipeline (preprocessing,
suggested detectors and segmenters, and parameter hints) while leaving
implementation to the agent. Both specify the exact output format. Every instruction forbids the agent
from retrieving the source publication or its reported values and from
identifying it through file names or metadata; general methods literature and
software documentation are allowed. Both instructions are reproduced in full for one task, together with the
shared footer, in Appendix~\ref{app:instructions}.

\paragraph{Required outputs.}\label{sec:outputs}
Each task declares its required output files with a filename pattern, a format
and, for tables, required columns (for example, one CSV of centroids per image
for counting, a Cell Tracking Challenge label stack plus lineage file for
tracking, an instance-label TIFF for segmentation). Predictions are matched to
ground truth by normalized filename stem or numeric index. The code, reports, figures and session transcript that the agent leaves alongside these files do not enter the outcome score and serve instead as the evidence for the process score.

\subsection{Agent integration}\label{sec:agents}

Each agent is integrated through a wrapper that passes the rendered task
instruction to the agent's native invocation and collects a standardized
submission directory. The agent receives a
read-only input directory and a writable output directory, runs until it stops
or the wall-clock budget elapses ($4$\,h; most model-exchange sessions used $2$\,h, Appendix~\ref{app:adapters}), and leaves its
output files in the output directory.
The wall-clock budget is enforced externally by running each task in an isolated subprocess and terminating its process group when the budget elapses. Within this limit, each agent retains its native per-step and per-tool policies, which are treated as part of the agent under test.
A run that reaches the wall-clock limit is scored on the files it has written before termination.
Because agents differ in how they define turns and tool calls, these counts are logged for
diagnostic purposes only.
We provide wrappers for six agents across three classes: an
LLM tool-use agent that writes and executes Python (Biomni~\citep{biomni});
three coding command-line agents (Claude Code, Codex and DeepSeek Harness),
each pointed at the study's models through OpenRouter's Anthropic- or
OpenAI-compatible endpoints, which lets the model behind a harness be
exchanged; and two GUI agents that operate Fiji/ImageJ
(CopilotJ~\citep{copilotj} and Agentic-J~\citep{agenticj}). No agent receives a system
prompt beyond the task instruction. Network access is not restricted, and every
agent except Codex exposes a web-search tool. An
audit of every archived trace found network access used only for package and
model-weight downloads and, in two runs, for software documentation.
Additional agents, such as napari- or model-zoo-based
systems~\citep{omega,gencellagent}, can be added through the same interface.
Per-agent invocation, token parsing and declared inner limits are given in
Appendix~\ref{app:adapters}.

\subsection{Scoring}\label{sec:scoring}
\paragraph{Outcome score.}\label{sec:outcome}
The outcome score is a task-specific metric computed against ground
truth and mapped to $[0,1]$: Dice for segmentation;
the Cell Tracking Challenge segmentation (SEG) and tracking (TRA) measures; F1 and localization
error for spot detection; the reproduced significance pattern and direction of condition differences, or the distribution of colocalization onset times, for colocalization; and distributional or rank-correctness
metrics (the Kolmogorov--Smirnov statistic or the relative error of a
derived quantity) for kinetics and feature extraction. The NF-$\kappa$B
translocation task, for which BBBC014 provides no per-well reference, is scored
on the dose--response properties of the submitted table (Appendix~\ref{app:gtreplay}). 
Every run is assigned one of four end states, delivered, no deliverable, crashed, and refused by the provider's safety filter (Table~\ref{edtab:profile}); runs that delivered but scored below $0.1$ are further split by whether
the closing message \textit{claimed} completion (Table~\ref{edtab:ladder}). A ground-truth replay reaches $\approx 1.0$ on every task
with a ground-truth file, and submissions without output files score $0$
(Appendix~\ref{app:gtreplay}).

\paragraph{Process score.}\label{sec:process}
The process score is a severity-weighted rubric over structured pipeline
sections (load~$\rightarrow$ preprocess~$\rightarrow$ segment/detect~$\rightarrow$ measure~$\rightarrow$ statistics~$\rightarrow$ visualize~$\rightarrow$ report),
assessing how the analysis was carried out and reported at every stage from
loading the raw data to the final report, independent of the final number.
The judge rates each rubric item pass, fail or unknown from the artifacts the
run saved (code, logs, figures, tables); the score is the
severity-weighted fraction of passed items among the items the judge could
decide. Items rated unknown are excluded from the denominator, as the evidence a run
preserves depends on the harness (e.g., a GUI agent leaves no code). The fraction of undecidable items is reported separately as an auditability measure. A run without the required output files may still receive a process score if it leaves code, figures or reports from which the judge can assess its analysis, and a run in which no item is decidable receives none. The process score is reported alongside the outcome score; rankings use
the outcome score alone.

\paragraph{Judge validation.}\label{sec:judge}
To validate the VLM judge, twenty runs, drawn to cover all six agents and all
$16$ tasks, were reviewed item by item by an expert cell biologist on a blinded
copy of each run folder in which every judge decision had been erased
($1{,}273$ rubric items; the expert decided $1{,}085$ and skipped $188$ as
undecidable from the folder or not applicable). On the $922$ items that both
decided, the adopted judge agreed with the expert on $87\%$ (Cohen's
$\kappa=0.67$, $95\%$ CI $0.61$--$0.73$), passing $27\%$ of the items the
expert marked as failed and failing $7\%$ of those the expert marked as passed, a net leniency of
two points in pass rate. Among the rubric subsections shown in Fig.~\ref{fig:ed-judge}b, agreement was highest for input understanding
($\kappa=0.87$) and lowest for tool choice ($\kappa=0.47$). Two further
candidate judges (Claude Opus~5, Gemini~3.1~Pro) scored the same runs and
matched the expert moderately well ($\kappa=0.66$ and $0.67$). Because judge
models carry model-specific biases~\citep{mtbench}, the benchmark fixes one
judge, Claude Sonnet~5, and reports its agreement with the expert alongside
every process score. At the run
level, process scores computed from the expert's labels correlated with the
judge's at $r=0.54$ and with the outcome score at $r=-0.03$ (Section~\ref{sec:res-process}). Together, the two scores provide complementary views of agent performance, with the process score reflecting interpretability and traceability and the outcome score measuring result correctness. The rubric,
the judge prompt and the full agreement tables are given in Appendix~\ref{app:judge}.

\section{Experimental setup}\label{sec:setup}

\subsection{Configurations}\label{sec:design}
\label{sec:harnessmodel}
Across $16$ tasks with three runs per configuration, the study separates the contribution of the harness from that of the model~\citep{clawswebench}. We first held the model constant and evaluated all six AI agents on GPT-5.6 Sol. We then held the harness constant and swapped between open- and closed-weight models on DeepSeek Harness (Opus 5, Kimi K2.6, GLM-5.1, V4-Pro and V4-Flash) and on Claude Code (Opus 5 and Kimi K2.6), so that DeepSeek Harness ran six models and Claude Code three. Lastly, we re-tested DeepSeek Harness on GPT-5.6 Sol and V4-Flash using expanded, detailed instructions. Certain pairings proved technically unviable. Specifically, two open-weight models, GLM-5.1 and V4-Flash, failed under the Claude Code harness because intermittent empty API responses were mistakenly logged as finished turns, and repeated testing confirmed this pattern. As a result, we omitted these incompatible setups from our scores. Similarly, nine runs of the wound-healing task by DeepSeek Harness on V4-Flash with the detailed instruction, which ended in a harness error because the request exceeded the provider's image-size limit for multi-image calls, are excluded and the task was rerun until three scored runs existed.

All figures and tables are regenerated directly from the run records. For any agent with a reasoning or ``thinking'' control, we fixed the effort to a single declared level and logged the exact setting in the run's manifest. If an agent lacked this setting, we noted that instead.
Runs that failed due to infrastructure issues, like container startup errors or scheduler preemption, were retried once. If they failed a second time for the same reason, they were excluded from the agent's overall statistics. However, agent-related failures such as crashes, timeouts, or empty submissions were never retried and were scored as observed.  
Every task ran on a single GPU with a read-only input mount. Each run's metadata records the GPU model, software versions, and model and judge identifiers (which are also listed in Table~\ref{tab:prices}). Finally, the code repository documents all job-generation commands and manifest fields.

\subsection{Measurement and statistics}\label{sec:measure}
\paragraph{Efficiency and cost.}\label{sec:cost}
Efficiency is reported as input and output token counts, wall-clock runtime,
tool-call counts and monetary cost per run. Cost has two
sources, stated per harness--model configuration: for the Claude Code configurations and
the DeepSeek-V4-Pro configuration it is the provider's own per-request
billing, summed over every request of every run; for all other
configurations it is the agent's metered token usage
priced at the provider's published per-token rates at the time of the study.
Cached input tokens are priced at the cache rate. Agents expose different usage
signals. For each agent we state which signals are measured and mark the rest as unavailable, so a missing count is never read as zero. Wall-clock runtime is the one measure
available for every agent. Agent token usage is accounted
separately from the judge's own token usage. Per-agent signal availability
and the token-counting convention are given in Appendix~\ref{app:usage}; the list
prices behind every derived cost are given in Table~\ref{tab:prices}.

\paragraph{Statistics.}\label{sec:stats}
Outcome scores are summarized per agent--task pair as the mean of its three
runs and per configuration as the mean over tasks, with the standard deviation (s.d.) across
tasks and the fraction of runs that delivered a required output file
(Tables~\ref{edtab:profile} and~\ref{edtab:ladder}). Outcome scores are also
stratified by subtask, dimensionality and difficulty level. Run-to-run variability
is reported as both the range of the outcome score across the three runs of each
agent--task pair and as the relative variance contributed by each experimental factor (Section~\ref{sec:res-variability}). Runtime, token, and
tool-call figures are reported as medians over scored runs, which are robust
to the heavy tail that budget-limited runs induce. 

The relation between
process and outcome scores is reported as a Pearson correlation over the scored
runs. Expert--judge agreement is reported as raw accuracy and Cohen's $\kappa$
over the items both decided, overall and per rubric stratum, with $95\%$
intervals from a cluster bootstrap that resamples whole runs ($10^4$
resamples); abstentions on either side are excluded from agreement and reported as rates. Runs blocked by
the provider's safety filter are excluded from outcome means; all other
runs, including crashes and timeouts, are retained and scored as observed. 
For the SARS-CoV-2 task, provider safety filters blocked some models from running. Codex was halted directly by a content-policy refusal message. Meanwhile, Claude Code’s three runs ended abruptly with empty responses from the API gateway after a few turns. Since this failure mode was absent from Claude Code’s other 141 runs, we categorized it as an API-level safety refusal.

\section{Results}\label{sec:results}

\subsection{Performance across tasks and agents}\label{sec:res-agents}
\begin{figure*}[p]
\centering
\includegraphics[width=\textwidth]{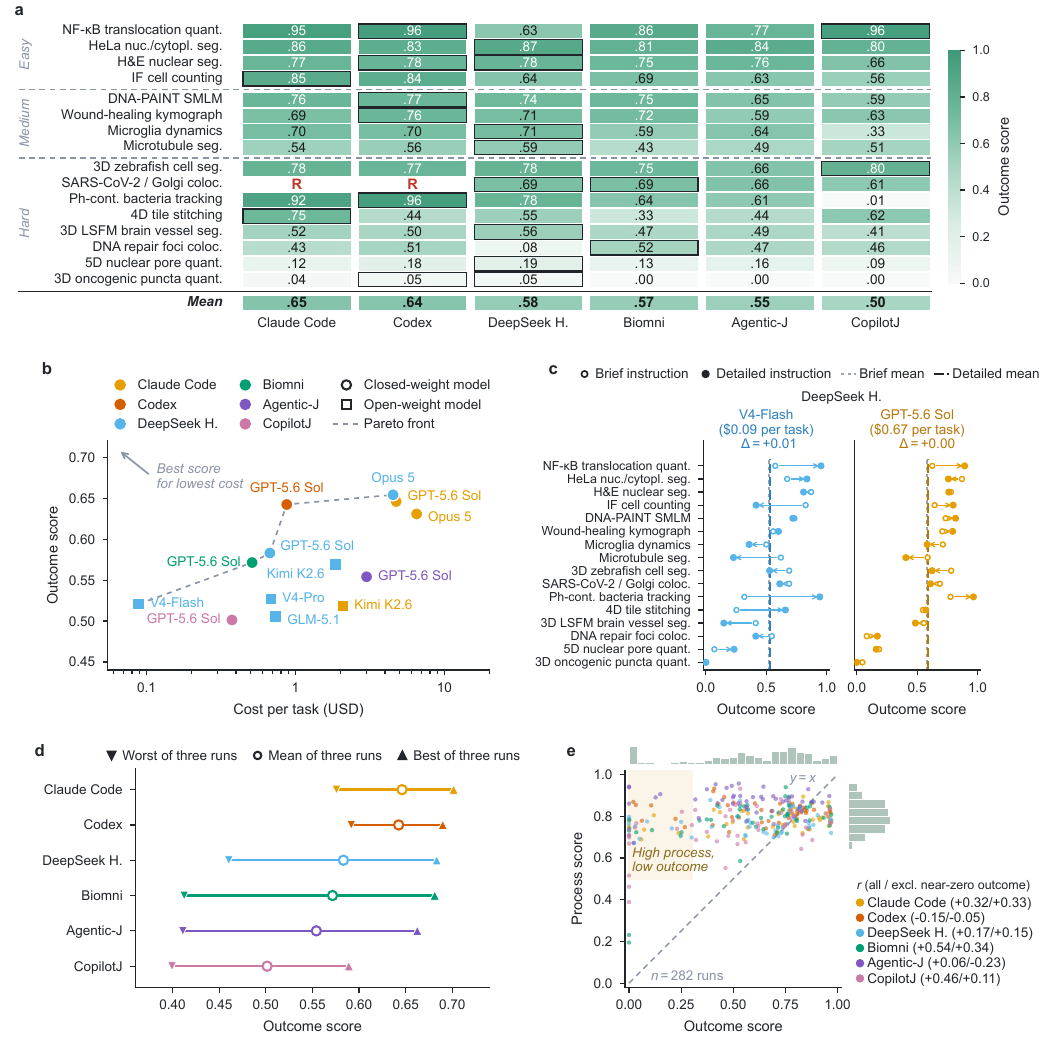}%
\caption{\textbf{Agent performance on \bench.}
\textbf{a)} Outcome score of the six agents, all driven by GPT-5.6~Sol, on each
of the $16$ tasks. Each entry is the mean of $n=3$ independent runs. Boxes mark the best agent
per task (ties boxed jointly) while rows are grouped by difficulty level. In the bottom row we computed the
mean over tasks per agent. Runs blocked by the model provider's safety filter are marked with R and are excluded from the means.
\textbf{b)} Mean outcome score against mean cost per task for $13$
harness--model configurations, the six agents on GPT-5.6~Sol and further models
on Claude Code and DeepSeek Harness (DeepSeek H.). The dashed line marks the Pareto front while circles and squares are the closed- and open-weight models respectively.
\textbf{c)} Effect of a detailed instruction. Mean outcome score per task
($n=3$ runs) under the brief instruction (open circles) and the detailed
instruction (filled circles) for DeepSeek Harness on V4-Flash (left) and GPT-5.6~Sol
(right). Arrows mark changes of at least $0.05$, the mean over tasks under each set of instructions is represented by a vertical line, and $\Delta$ is the difference between the two means.
\textbf{d)} Run-to-run variability. The mean outcome score for each agent over all tasks ($n=16$) is marked by open
circles while the worst and best scores per task are represented by downward and upward triangles, respectively.
\textbf{e)} Process score ($y$-axis) plotted against outcome score ($x$-axis) for all scored runs ($n=282$), with colors corresponding to each agent. Pearson's $r$ per agent is provided across all runs as well as for runs with an outcome above $0.05$. Marginal histograms display the distribution of each individual score.  The shaded region highlights runs that demonstrated a sound process (process score $\ge 0.5$) yet resulted in a low outcome ($\le 0.3$), while the dashed line represents $y=x$.}
\label{fig:findings}
\end{figure*}

\begin{figure}[tbp]
\centering
\includegraphics[width=\textwidth]{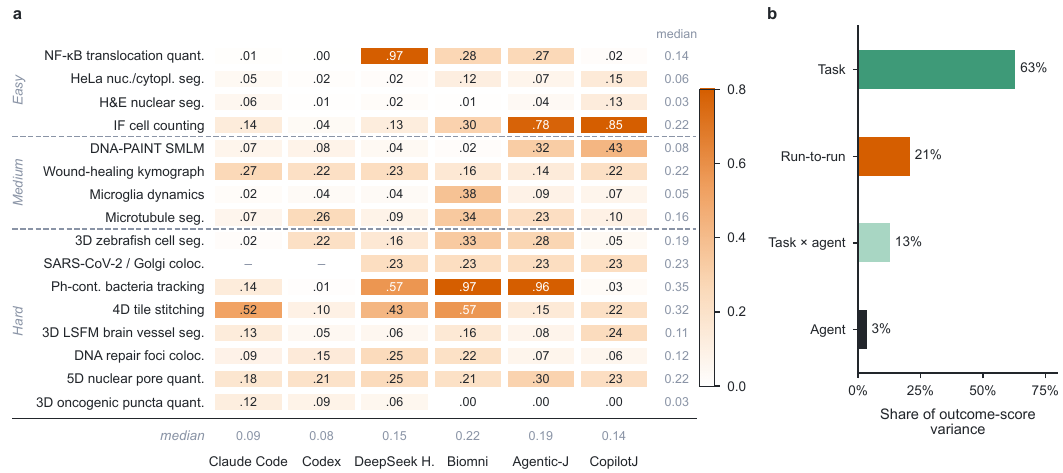}
\caption{\textbf{Run-to-run variability of the outcome score.} \textbf{a)} For every
agent--task pair of the GPT-5.6~Sol study, the range of the outcome score across
the pair's three independent runs (rows and columns ordered as in Fig.~\ref{fig:findings}a;
gray margins, medians; dashes, pairs with fewer than two scored runs (safety-filter
refusals)). Median range, $0.08$ to
$0.22$ across agents and $0.03$ to $0.35$ across tasks. In $40\%$ of pairs
the same agent moved by more than $0.2$ between identical runs.
\textbf{b)} 
Breakdown of total outcome-score variance across all scored runs, showing the relative contributions of the task, task–agent interactions, run-to-run instability within a pair, and the choice of agent. Notably, run-to-run variability accounts for six times as much variance as the agent chosen.}
\label{fig:ed-reliability}
\end{figure}

\begin{figure}[tbp]
\centering
\includegraphics[width=\textwidth]{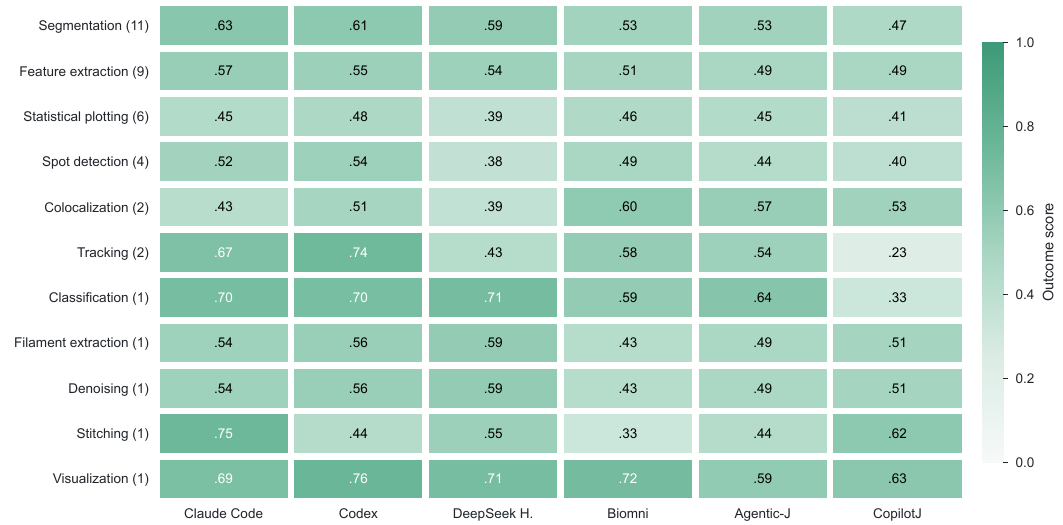}
\caption{\textbf{Outcome score by analysis subtask.} 
Mean outcome score for each agent (using GPT-5.6~Sol), calculated across all tasks involving a given subtask type. The number of relevant tasks appears in parentheses, with individual tasks contributing to all subtasks they include (rows are not mutually exclusive). Values represent averages of the per-task means across three independent runs, excluding any attempts blocked by provider safety filters.}
\label{fig:ed-capability}
\end{figure}

To isolate the effect of agent design, we first evaluated six agents using the same language model (i.e., GPT-5.6~Sol), with three independent runs per agent and task (Table~\ref{edtab:profile}). Three were general-purpose coding agents (Claude Code, Codex and DeepSeek Harness), whereas the other three were designed for biology (Biomni, CopilotJ and Agentic-J). Outcome scores varied substantially more across tasks than across agents (Fig.~\ref{fig:findings}a; Fig.~\ref{fig:ed-reliability}b). On NF-$\kappa$B translocation quantification, HeLa nucleus and cytoplasm segmentation and cell counting, the best agents scored $0.85$--$0.96$ and every agent reached at least $0.56$. However, tasks that added a third dimension or a time axis included the hardest ones, on which scores dropped to $0.19$ for nuclear-pore assembly kinetics and $0.05$ for 3D puncta quantification (Table~\ref{edtab:ladder}). Additionally, provider safety filters blocked Claude Code and Codex on the SARS-CoV-2 Golgi colocalization task (Fig.~\ref{fig:findings}a, R; Table~\ref{edtab:profile}). Biological specialization conferred no overall advantage across these tasks. General-purpose agents achieved or tied the highest mean outcome on $14$ of $16$ tasks, outperforming biology-specific agents by up to $0.32$ on bacterial tracking (Fig.~\ref{fig:ed-capability}), and biology-specific agents showed no advantage on the five three-dimensional tasks either (differences of $-0.13$ to $+0.02$). Given that some bioimaging-specialized agents were designed primarily for human-in-the-loop use, this result is expected when they run autonomously.

\begin{table}[htbp]\centering
\caption{\textbf{Per-configuration profile.} Outcome is the mean over per-task means (three runs per task), $\pm$ the standard deviation (s.d.) across tasks and process is the mean process score. The D/N/C/R column reports respectively the number of runs which delivered a required output file, ended without one, crashed, or were refused by the provider's safety filter. Time and tokens are medians over scored runs. Cost is reported in US Dollars (USD, \$) per run. Asterisks (\textsuperscript{*}) mark values measured from per-request billing records, while all others are calculated from native token usage at list price. Claude Code, Codex and DeepSeek H. are general-purpose coding agents, while Biomni, Agentic-J and CopilotJ are biology-specific.}\label{edtab:profile}
{\small\setlength{\tabcolsep}{2pt}
\begin{tabular*}{\textwidth}{@{\extracolsep{\fill}}lccccccr@{\extracolsep{0pt}}l@{}}
\toprule
Harness & Model & Outcome $\pm$ s.d. & Process & D/N/C/R & Time (min) & Tokens in/out ($\times10^{3}$) & \multicolumn{2}{c}{\$/run} \\
\midrule
Claude Code & GPT-5.6 Sol & 0.65 $\pm$ 0.27 & 0.79 & 45/0/0/3 & 11.4 & 994/10.7 & 4.75 & \textsuperscript{*} \\
Codex & GPT-5.6 Sol & 0.64 $\pm$ 0.27 & 0.82 & 45/0/0/3 & 10.3 & 2156/19.6 & 0.87 &  \\
DeepSeek H. & GPT-5.6 Sol & 0.58 $\pm$ 0.25 & 0.77 & 47/1/0/0 & 10.4 & 1245/14.4 & 0.67 &  \\
Biomni & GPT-5.6 Sol & 0.57 $\pm$ 0.24 & 0.77 & 46/2/0/0 & 7.4 & 270/15.4 & 0.51 &  \\
Agentic-J & GPT-5.6 Sol & 0.55 $\pm$ 0.22 & 0.87 & 47/1/0/0 & 31.7 & 2946/88.2 & 3.01 &  \\
CopilotJ & GPT-5.6 Sol & 0.50 $\pm$ 0.28 & 0.74 & 48/0/0/0 & 7.3 & 245/18.1 & 0.38 &  \\
\midrule
Claude Code & Kimi K2.6 & 0.52 $\pm$ 0.31 & 0.70 & 41/7/0/0 & 42.1 & 3120/45.9 & 2.09 & \textsuperscript{*} \\
Claude Code & Opus 5 & 0.63 $\pm$ 0.24 & 0.89 & 48/0/0/0 & 27.5 & 4152/51.6 & 6.52 & \textsuperscript{*} \\
DeepSeek H. & GLM-5.1 & 0.51 $\pm$ 0.22 & 0.73 & 45/3/0/0 & 37.1 & 2041/46.2 & 0.74 &  \\
DeepSeek H. & Kimi K2.6 & 0.57 $\pm$ 0.25 & 0.72 & 45/3/0/0 & 40.0 & 2866/49.2 & 1.85 &  \\
DeepSeek H. & Opus 5 & 0.65 $\pm$ 0.26 & 0.85 & 46/1/0/0 & 22.0 & 2556/45.1 & 4.53 &  \\
DeepSeek H. & V4-Flash & 0.52 $\pm$ 0.25 & 0.72 & 44/4/0/0 & 32.0 & 2389/48.9 & 0.09 &  \\
DeepSeek H. & V4-Pro & 0.53 $\pm$ 0.27 & 0.71 & 48/0/0/0 & 38.4 & 2439/50.1 & 0.69 & \textsuperscript{*} \\
\bottomrule
\end{tabular*}}
\end{table}

\begin{table}[htbp]\centering
\caption{\textbf{Capability indicators per configuration.} Metrics are grouped into five categories containing two indicators each. \emph{Deliver}: percentage of runs that reached scoring with deliverables, and percentage stopped by the scheduler at the wall-clock limit. \emph{Understand}: rubric pass rate for input understanding (Input) and tool choice and use (Tools). \emph{Quantify}: rubric pass rate for the segmentation stage (Segm.) and, pooled, for the measurement stages (quantification, feature extraction, statistics and plotting, tracking, colocalization, spot detection; Quant.). \emph{Dimensionality}: mean outcome on the eleven 2D and time-lapse tasks (2D) and on the five 3D and 5D tasks (3D/5D). \emph{Process vs outcome}: Pearson $r$ between process and outcome score over runs with outcome $>0.05$, and, among runs that delivered but scored $<0.1$, how many closed with a message claiming completion (Claims). A dash (--) denotes that no delivered run scored $<0.1$. Rubric rates count decided items only. For Agentic-J and CopilotJ the closing message is the harness's own summary, for the others the model's final turn. Rows show the six agents on GPT-5.6 Sol, the model exchanges, and the detailed-instruction study.}\label{edtab:ladder}
{\small\setlength{\tabcolsep}{0.6pt}
\begin{tabular*}{\textwidth}{@{\extracolsep{\fill}}l@{\hspace{7pt}}lcccccccccc@{}}
\toprule
 & & \multicolumn{2}{c}{Deliver} & \multicolumn{2}{c}{Understand} & \multicolumn{2}{c}{Quantify} & \multicolumn{2}{c}{Dimensionality} & \multicolumn{2}{c}{Process vs outcome} \\
\cmidrule(lr){3-4}\cmidrule(lr){5-6}\cmidrule(lr){7-8}\cmidrule(lr){9-10}\cmidrule(lr){11-12}
Harness & Model & Delivered & Timed out & Input & Tools & Segm. & Quant. & 2D & 3D/5D & $r$ & Claims \\
\midrule
Claude Code & GPT-5.6 Sol & 100 & 0 & 0.89 & 0.76 & 0.85 & 0.67 & 0.75 & 0.44 & 0.33 & 4/4 \\
Codex & GPT-5.6 Sol & 100 & 0 & 0.96 & 0.94 & 0.85 & 0.69 & 0.77 & 0.39 & -0.05 & 3/3 \\
DeepSeek H. & GPT-5.6 Sol & 98 & 0 & 0.89 & 0.63 & 0.83 & 0.68 & 0.66 & 0.42 & 0.15 & -- \\
Biomni & GPT-5.6 Sol & 96 & 0 & 0.93 & 0.76 & 0.77 & 0.64 & 0.68 & 0.34 & 0.34 & 3/4 \\
Agentic-J & GPT-5.6 Sol & 98 & 4 & 0.94 & 0.74 & 0.83 & 0.83 & 0.65 & 0.35 & -0.23 & 0/3 \\
CopilotJ & GPT-5.6 Sol & 100 & 2 & 0.93 & 0.67 & 0.73 & 0.64 & 0.56 & 0.38 & 0.11 & 0/8 \\
\midrule
Claude Code & Kimi K2.6 & 85 & 21 & 0.89 & 0.68 & 0.76 & 0.55 & 0.62 & 0.30 & 0.58 & 2/3 \\
Claude Code & Opus 5 & 100 & 4 & 0.90 & 0.88 & 0.88 & 0.84 & 0.71 & 0.45 & -0.12 & 3/3 \\
DeepSeek H. & GLM-5.1 & 94 & 19 & 0.90 & 0.57 & 0.75 & 0.59 & 0.57 & 0.37 & 0.20 & -- \\
DeepSeek H. & Kimi K2.6 & 94 & 21 & 0.89 & 0.63 & 0.70 & 0.57 & 0.65 & 0.40 & 0.44 & -- \\
DeepSeek H. & Opus 5 & 98 & 6 & 0.90 & 0.79 & 0.86 & 0.80 & 0.75 & 0.44 & 0.13 & -- \\
DeepSeek H. & V4-Flash & 92 & 2 & 0.89 & 0.58 & 0.73 & 0.56 & 0.63 & 0.28 & 0.04 & -- \\
DeepSeek H. & V4-Pro & 100 & 0 & 0.88 & 0.55 & 0.72 & 0.56 & 0.62 & 0.32 & 0.27 & 9/9 \\
\midrule
DeepSeek H. & V4-Flash (detailed) & 100 & 2 & 0.89 & 0.48 & 0.78 & 0.60 & 0.63 & 0.31 & 0.36 & 4/4 \\
DeepSeek H. & GPT-5.6 Sol (detailed) & 100 & 0 & 0.89 & 0.56 & 0.85 & 0.70 & 0.69 & 0.37 & -0.19 & 4/6 \\
\bottomrule
\end{tabular*}}
\end{table}


\subsection{Models, cost and instructions}\label{sec:res-model}
Because the preceding evaluation held the language model constant, agent differences reflected the harness alone. We then asked how much the model contributes compared to the harness. Six harnesses on GPT-5.6~Sol had mean outcomes from $0.50$ to $0.65$, and six models on DeepSeek Harness spanned almost the same range (Fig.~\ref{fig:findings}b and Table~\ref{edtab:matrix}). Opus~5 improved DeepSeek Harness by $0.07$ but barely changed Claude Code, underscoring the importance of harness--model alignment, as a harness may be designed primarily for specific models. The most economical configuration, DeepSeek-V4-Flash, reached $80\%$ of the strongest configuration's score for $2\%$ of its cost, while moving up the Pareto front from Codex cost five times as much for a gain of only $0.01$. This shows that more expensive configurations were not necessarily more accurate.

\begin{table}[htbp]\centering
\caption{\textbf{Outcome score per task for the model exchanges of Fig.~\ref{fig:findings}b, averaged over three runs.} Columns indicate the model behind Claude Code or DeepSeek Harness (DeepSeek H.), whereas the six agents on GPT-5.6~Sol are given in Fig.~\ref{fig:findings}a. Rows are grouped by difficulty level and ordered within a level as in Fig.~\ref{fig:findings}a. GLM-5.1 and V4-Flash could not run behind Claude Code (Section~\ref{sec:harnessmodel}). The dagger symbol (\dag) denotes fewer than three scored runs. The bottom row displays the mean across all tasks. LSFM, light-sheet fluorescence microscopy; IF, immunofluorescence.}\label{edtab:matrix}
{\small\setlength{\tabcolsep}{2pt}
\begin{tabular*}{\textwidth}{@{\extracolsep{\fill}}clccccccc@{}}
\toprule
 & & \multicolumn{2}{c}{Claude Code} & \multicolumn{5}{c}{DeepSeek H.} \\
\cmidrule(lr){3-4}\cmidrule(lr){5-9}
 & Task & Kimi K2.6 & Opus 5 & GLM-5.1 & Kimi K2.6 & Opus 5 & V4-Flash & V4-Pro \\
\midrule
\multirow{4}{*}{\rotatebox{90}{Easy}} & NF-$\kappa$B translocation quant. & 0.85 & 0.97 & 0.87 & 0.94 & 0.98 & 0.57 & 0.72 \\
 & HeLa nuc./cytopl. seg. & 0.87 & 0.61 & 0.71 & 0.69 & 0.88 & 0.67 & 0.75 \\
 & H\&E nuclear seg. & 0.87 & 0.88 & 0.71 & 0.83 & 0.88 & 0.87 & 0.85 \\
 & IF cell counting & 0.40 & 0.59 & 0.43 & 0.48 & 0.75\dag & 0.83 & 0.39 \\
\midrule
\multirow{4}{*}{\rotatebox{90}{Medium}} & DNA-PAINT SMLM & 0.78 & 0.67 & 0.46 & 0.75 & 0.76 & 0.73 & 0.70 \\
 & Wound-healing kymograph & 0.69 & 0.63 & 0.47 & 0.79 & 0.75 & 0.56 & 0.78 \\
 & Microglia dynamics & 0.46 & 0.72 & 0.48 & 0.34 & 0.72 & 0.50 & 0.51 \\
 & Microtubule seg. & 0.60 & 0.59 & 0.33 & 0.60 & 0.64 & 0.62 & 0.29 \\
\midrule
\multirow{8}{*}{\rotatebox{90}{Hard}} & 3D zebrafish cell seg. & 0.33 & 0.84 & 0.58 & 0.51 & 0.84 & 0.69 & 0.58 \\
 & SARS-CoV-2 / Golgi coloc. & 0.77 & 0.77 & 0.77 & 0.77 & 0.69 & 0.69 & 0.77 \\
 & Ph-cont. bacteria tracking & 0.00 & 0.97 & 0.60 & 0.64 & 0.97 & 0.32 & 0.65 \\
 & 4D tile stitching & 0.84 & 0.48 & 0.61 & 0.70 & 0.49 & 0.25 & 0.75 \\
 & 3D LSFM brain vessel seg. & 0.17 & 0.58 & 0.50 & 0.52 & 0.53 & 0.41 & 0.25 \\
 & DNA repair foci coloc. & 0.48 & 0.48 & 0.43 & 0.30 & 0.26 & 0.55 & 0.46 \\
 & 5D nuclear pore quant. & 0.17 & 0.23 & 0.13 & 0.18 & 0.25 & 0.07 & 0.00 \\
 & 3D oncogenic puncta quant. & 0.01 & 0.09 & 0.01 & 0.06 & 0.10 & 0.00 & 0.00 \\
\midrule
 & \emph{Mean} & 0.52 & 0.63 & 0.51 & 0.57 & 0.65 & 0.52 & 0.53 \\
\bottomrule
\end{tabular*}}
\end{table}

\label{sec:res-levers}
We next investigated whether failure on complex tasks could be rescued by stronger models or more detailed instructions. On DeepSeek Harness, replacing V4-Flash with GPT-5.6~Sol increased the mean outcome from $0.52$ to $0.58$, and Opus~5 reached $0.65$ (Fig.~\ref{fig:findings}b). By contrast, replacing the brief instruction with a detailed protocol that a bioimage-analysis expert derived from the source study left the mean almost unchanged ($+0.01$ on V4-Flash and $<0.01$ on GPT-5.6~Sol), despite mean absolute task-level changes of $0.21$ and $0.11$, respectively (Fig.~\ref{fig:findings}c). On V4-Flash, the detailed instruction raised bacterial tracking from $0.32$ to $0.94$ but lowered cell counting from $0.83$ to $0.42$. No combination of model and instruction exceeded $0.25$ on nuclear-pore assembly kinetics or 3D puncta quantification. More detailed instructions therefore redistributed performance rather than improving it overall, and neither change rescued the tasks that every agent failed.

\subsection{Run-to-run variability and indicators of correctness}\label{sec:res-variability}
Scores also varied from run to run, due to the nature of randomness of the underlying models. Across three attempts on the same task, they differed by more than $0.2$ in $40\%$ of agent--task pairs, and taking the best of three brought five of the six agents within $0.04$ of one another, whereas their worst attempts differed by $0.18$ (Fig.~\ref{fig:findings}d), indicating that differences between agents primarily reflected reliability rather than best-case capability.

\begin{figure}[tbp]
\centering
\includegraphics[width=\textwidth]{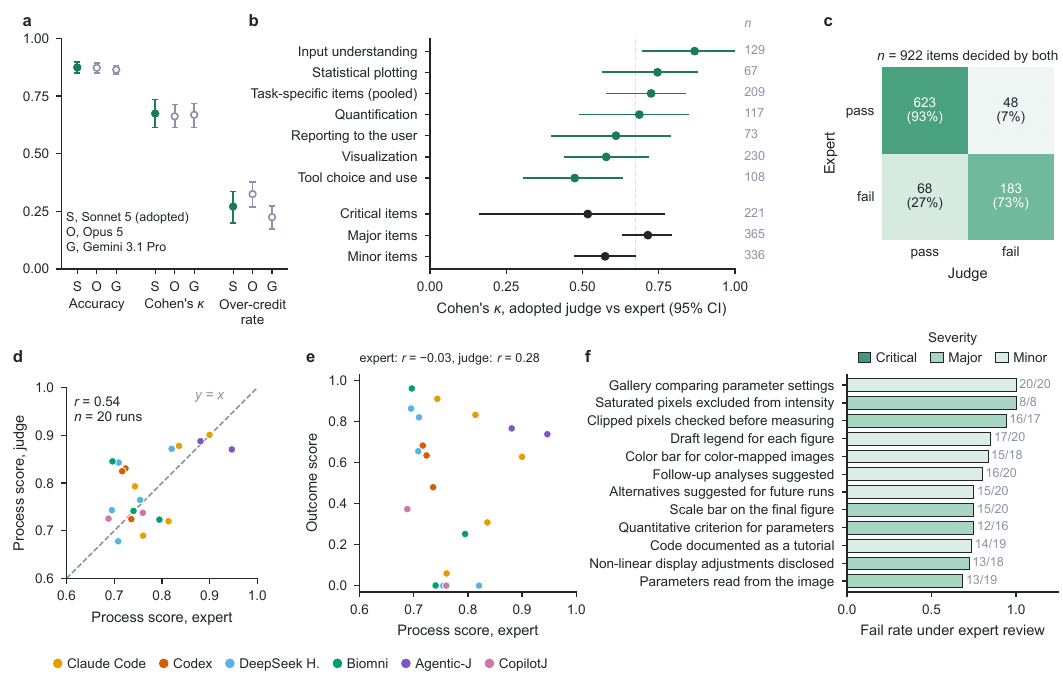}
\caption{\textbf{Agreement between the vision--language judge and a human expert.} Twenty runs (six agents on GPT-5.6~Sol, covering all $16$ tasks) were reviewed item by item by an expert cell biologist blinded to the judge's decisions ($1{,}273$ total rubric items, with $188$ skipped as undecidable from the run folder or not applicable). 
\textbf{a)} Agreement with the expert across three candidate VLM judges on items decided by both, showing accuracy, Cohen's $\kappa$, and the over-crediting rate (cases where the judge assigned a pass to an item marked as failed by the expert). Error bars denote $95\%$ cluster-bootstrap intervals across runs. Finally, Sonnet~5 is the adopted judge in the rest of the reported findings. 
\textbf{b)} Agreement ($\kappa$) of the adopted judge stratified by rubric subsection (green) and item severity (black). The grayed $n$ values represent the number of items evaluated by both the human expert and the VLM judge. The dashed vertical line marks the overall average $\kappa$. 
\textbf{c)} Confusion matrix of the adopted VLM judge evaluated against the human expert with each cell reporting raw item counts and percentages normalized across the expert's true classes.
\textbf{d)} Per-run process score under expert labels against under the judge, each computed using the rubric's severity weights across decided items. The dashed line represents $y=x$. 
\textbf{e)} The expert's process score plotted against the outcome score for the same runs ($r=-0.03$), where the five runs with an outcome below $0.1$ still received expert process scores of $0.74$--$0.82$. 
\textbf{f)} The twelve rubric items failed most frequently under expert review (among items evaluated in at least eight runs), colored by severity, with fractions indicating failed over decided runs. In total, runs failed $4\%$ of critical items, $31\%$ of major items, and $44\%$ of minor items under expert review. Marker colors in \textbf{d)} and \textbf{e)} identify agents as defined in the bottom legend.}
\label{fig:ed-judge}
\end{figure}

\begin{figure}[tbp]
\centering
\includegraphics[width=\textwidth]{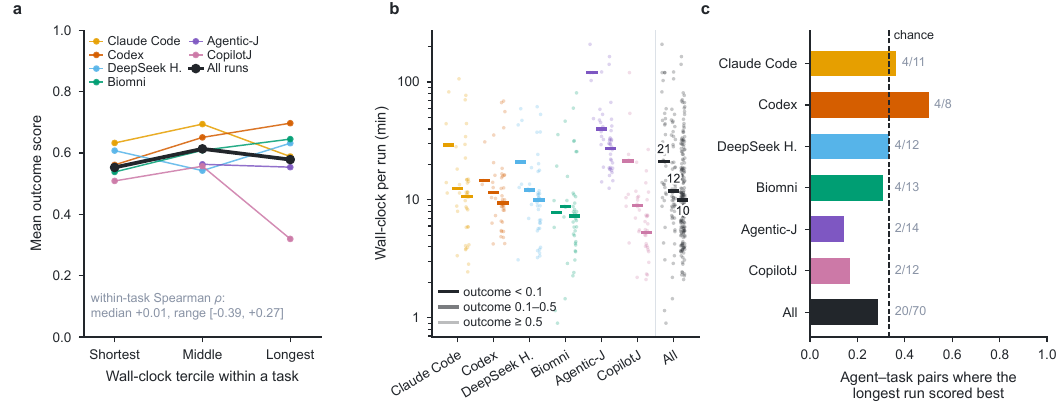}
\caption{\textbf{Wall-clock time and outcome.} Data represent all scored runs from the GPT-5.6 Sol evaluation, where wall-clock time serves as the primary metric directly comparable across all agents (output tokens show the same pattern). \textbf{a)} Mean outcome score by wall-clock tercile. Within each task,
its scored runs (up to $18$, six agents with three runs each) were ranked by wall-clock time
and divided into the shortest, middle and longest third, plotted both pooled across all agents (black) and individually by agent (colors). The pooled mean differs by $0.03$ between the shortest and the
longest third and by at most $0.06$ between any two thirds, and within a
task the rank correlation between wall-clock time and outcome is centered
on zero (annotation). \textbf{b)} Wall-clock per run grouped by
agent and by outcome tier ($<0.1$, $0.1$--$0.5$, $\ge 0.5$, displayed left to right for each agent). Horizontal ticks indicate medians, and the rightmost column groups all agents. Runs with an outcome below $0.1$ took twice as long as those with an outcome of at least $0.5$ (median of $21$ versus $10$ minutes) and were slower for each of the six
agents. \textbf{c)} Fraction of agent–task pairs in which the longest run achieved the highest score, evaluated across the 70 pairs whose runs differed by more than 0.05. The vertical dashed line indicates expected performance by chance ($1/3$).}
\label{fig:ed-token}
\end{figure}

\label{sec:res-process}
Since the same agent could succeed on one run and fail on the next, we finally asked whether a run's success could be told, without ground truth, from its process score or wall-clock time. Across all scored runs, process and outcome scores correlated weakly (Pearson $r=0.29$; Fig.~\ref{fig:findings}e), and excluding runs with near-zero outcomes reduced $r$ to $0.09$. The judge also compressed its ratings into a narrow range, assigning almost no run a process score below $0.5$ (Fig.~\ref{fig:findings}e, right margin). Process scores assigned by a human expert to $20$ runs were no more predictive of outcome ($r=-0.03$; Fig.~\ref{fig:ed-judge}). When three runs of an agent--task pair disagreed, the longest run was the best in only $20$ of $70$ pairs, and runs that scored below $0.1$ took twice as long as those that succeeded (Fig.~\ref{fig:ed-token} and Table~\ref{edtab:time}). Thus, neither the process nor the time spent on an analysis can be used to indicate whether its result is scientifically correct (Appendix~\ref{app:vignettes} examines three such runs).

\begin{table}[htbp]\centering
\caption{\textbf{Wall-clock minutes per task for the six agents on GPT-5.6~Sol, median over three runs.} Rows are grouped and ordered as in Table~\ref{edtab:matrix}, and the bottom row displays the median over all scored runs of the agent, as in Table~1. The double dagger symbol (\ddag) denotes that at least one of the three runs was stopped at the wall-clock limit (Section~\ref{sec:stats}) and the median is right-censored.}\label{edtab:time}
{\small\setlength{\tabcolsep}{6pt}
\begin{tabular}{clcccccc}
\toprule
 & Task & Claude Code & Codex & DeepSeek H. & Biomni & Agentic-J & CopilotJ \\
\midrule
\multirow{4}{*}{\rotatebox{90}{Easy}} & NF-$\kappa$B translocation quant. & 12 & 10 & 17 & 7 & 27 & 10 \\
 & HeLa nuc./cytopl. seg. & 12 & 16 & 10 & 11 & 19 & 7 \\
 & H\&E nuclear seg. & 4 & 5 & 4 & 5 & 22 & 5 \\
 & IF cell counting & 11 & 9 & 10 & 9 & 33 & 5 \\
\midrule
\multirow{4}{*}{\rotatebox{90}{Medium}} & DNA-PAINT SMLM & 4 & 5 & 4 & 3 & 18 & 4 \\
 & Wound-healing kymograph & 11 & 10 & 10 & 7 & 30 & 6 \\
 & Microglia dynamics & 10 & 9 & 9 & 7 & 32 & 10 \\
 & Microtubule seg. & 4 & 5 & 4 & 2 & 15 & 2 \\
\midrule
\multirow{8}{*}{\rotatebox{90}{Hard}} & 3D zebrafish cell seg. & 23 & 17 & 24 & 31 & 34 & 14 \\
 & SARS-CoV-2 / Golgi coloc. & 3 & 6 & 6 & 4 & 18 & 4 \\
 & Ph-cont. bacteria tracking & 44 & 56 & 37 & 32 & 141 & 21\textsuperscript{\ddag} \\
 & 4D tile stitching & 9 & 9 & 11 & 6 & 48 & 7 \\
 & 3D LSFM brain vessel seg. & 14 & 10 & 11 & 9 & 45 & 7 \\
 & DNA repair foci coloc. & 18 & 16 & 14 & 10 & 42 & 9 \\
 & 5D nuclear pore quant. & 11 & 15 & 12 & 16 & 49 & 12 \\
 & 3D oncogenic puncta quant. & 82 & 21 & 47 & 19 & 120\textsuperscript{\ddag} & 22 \\
\midrule
 & \emph{All tasks} & 11 & 10 & 10 & 7 & 32 & 7 \\
\bottomrule
\end{tabular}}
\end{table}

\FloatBarrier

\section{Discussion}\label{sec:discussion}

In summary, an analysis that appears plausible or methodologically sound may still produce wrong results. Current agents can complete routine bioimage analyses but remain unreliable as tasks become more complex, still far away from human-level performance (because every task is scored at the study's own endpoint, \bench assumes human experts could achieve nearly perfect scores on these tasks in theory). Neither biological specialization, stronger models nor detailed expert instructions can yet reliably close this gap. Our results locate this gap in data complexity, reliability and self-checking. Some tasks that added a third dimension or a time axis stayed out of reach for every agent. Taking the best of three runs brought five of the six agents close to one another, yet the failing runs rarely looked like failures, delivering tables that contradicted their own reports, applying detection thresholds never checked against the images or closing with a promise to finish, and their process score and runtime did not set them apart. This work not only provides an evaluation benchmark, publicly hosted on Hugging Face, but also offers the AI community a general development strategy for building large-scale multistep image-to-insight datasets for bioimage analysis. We envision the long-term goal of expert-level autonomous bioimage analysis being achieved by the community in a few years.

\section*{Acknowledgements}
J.C., L.J. and Y.Z. were partially supported by Federal Ministry of Research, Technology and Space (Bundesministerium für Forschung, Technologie und Raumfahrt, BMFTR) under the funding reference 161L0272. D.P. was partially supported by NFDI4Bioimage, funded by the German Research Foundation (DFG) within the framework of the NFDI-project number 501864659. The work at ISAS was additionally supported by the ``Ministerium für Kultur und Wissenschaft des Landes Nordrhein-Westfalen'' and ``Der Regierende Bürgermeister von Berlin, Senatskanzlei Wissenschaft und Forschung'' in Germany. The work at Institute of Computer Science, University of Tartu, was conducted using the research infrastructure ``ELIXIR Estonia'' funded by the Estonian Research Council (TARISTU24-TK4).

\section*{Author contributions}
J.C. and Y.S. conceptualized the study. Z.P. and D.P. designed the method. D.P. collected and curated the data. D.P., Z.P., L.J., M.M. and Y.Z. contributed to the experimental design and evaluation. Z.P. implemented the experiments. Z.P. and D.P. drafted the paper. All authors reviewed, revised and approved the final version of the paper.

\section*{Competing interests}
The authors co-developed Agentic-J~\citep{agenticj}.

\section*{Data and code availability}
The task inputs and ground truth that support the findings of this study are
available in the Hugging Face repository \texttt{BIABench/BIABench} at
\url{https://doi.org/10.57967/hf/10561} (revision \texttt{v1.0}). All $16$
tasks are built from publicly available datasets, whose source repositories and
references are listed in Appendix~\ref{app:tasks}.
\bench, including the task specifications, evaluation code, rubrics, judge
prompt, agent wrappers and analysis scripts, is available at
\url{https://github.com/BIABench/BIABench} under the BSD 3-Clause license.

\bibliographystyle{unsrtnat}
\bibliography{sn-bibliography}

\startappendix
\appendixcontents
\FloatBarrier
\section{Task specifications and provenance}\label{note:1}\label{app:tasks}
Every task is rebuilt from a published biological study and inherits that
study's question and its ground truth. Table~\ref{tab:tasks}
lays the suite out as a task-by-subtask matrix, and
Table~\ref{tab:taskspec} gives the required output files and the
scoring metric of each task. The input data supplied to the agent range from $4$~MB (H\&E nuclear
segmentation) to $5.1$~GB (3D oncogenic puncta quantification), and the five
largest inputs are all three-dimensional or time-lapse data
(Table~\ref{tab:tasks}). The descriptions that follow are grouped by difficulty level
(easy, medium, hard); each opens with the shortened task label used in the
figures and tables, then names the source study, the biological target and what
the agent is asked to measure; the per-task
source accessions and the scripts that reconstruct the data locally are
included in the benchmark repository.

\begin{sidewaystable}[p]
\centering
\caption{\textbf{The \bench task suite as a subtask matrix.} The $16$ tasks
(derived from each \texttt{task\_spec.yaml}) are grouped by
\emph{Difficulty} level (first column); \emph{Task} is the shortened task label
defined in Appendix~\ref{app:tasks}, \emph{Modality} the imaging modality,
\emph{Dimension}/\emph{Temporal} report spatial dimensionality (2D/3D) and static
vs.\ time-lapse acquisition, and \emph{Data} is the size of the input data
supplied to the agent (images and metadata, in MB). The eleven
rightmost columns are the analysis subtasks (S1, segmentation; S2, feature
extraction; S3, statistical plotting; S4, spot detection; S5, colocalization;
S6, tracking; S7, classification; S8, filament extraction; S9, denoising; S10,
stitching; S11, visualization), and a bullet ($\bullet$) marks every subtask a
task exercises.
The difficulty level is assigned by hand. Easy tasks are single-subtask or a standard two-dimensional pipeline. Medium tasks stay two-dimensional but call for a specialized method, such as single-molecule localization, optical flow or filament extraction. Most Hard tasks combine several subtasks, and a few remain hard for a single but demanding subtask (e.g., single-cell tracking across an $800$-frame time-lapse).  DIC,
differential interference contrast; TIRF, total internal reflection
fluorescence; IF, immunofluorescence.}\label{tab:tasks}
{\small%
\setlength{\tabcolsep}{1.8pt}%
\renewcommand{\arraystretch}{1.2}%
\begin{tabular}{@{}cllccN{0000}*{11}{c}@{}}
\toprule
Difficulty & Task & Modality & Dimension & Temporal & \multicolumn{1}{c}{Data (MB)} & S1 & S2 & S3 & S4 & S5 & S6 & S7 & S8 & S9 & S10 & S11 \\
\midrule
\multirow{4}{*}{Easy} & NF-$\kappa$B translocation quant. & Fluorescence widefield & 2D & Static & 395 & $\bullet$ & $\bullet$ & $\bullet$ & & & & & & & & \\
       & HeLa nuc./cytopl. seg. & Fluorescence widefield & 2D & Static & 288 & $\bullet$ & & & & & & & & & & \\
       & H\&E nuclear seg. & Brightfield & 2D & Static & 4 & $\bullet$ & & & & & & & & & & \\
       & IF cell counting & Fluorescence widefield & 2D & Static & 40 & & & & $\bullet$ & & & & & & & \\
\midrule
\multirow{4}{*}{Medium} & DNA-PAINT SMLM & TIRF / SMLM & 2D & Static & 12 & & & & $\bullet$ & & & & & & & \\
       & Wound-healing kymograph & DIC & 2D & Time-lapse & 570 & & $\bullet$ & & & & & & & & & $\bullet$ \\
       & Microglia dynamics & Phase contrast & 2D & Time-lapse & 166 & $\bullet$ & $\bullet$ & $\bullet$ & & & & $\bullet$ & & & & \\
       & Microtubule seg. & Fluorescence & 2D & Static & 7 & $\bullet$ & & & & & & & $\bullet$ & $\bullet$ & & \\
\midrule
\multirow{8}{*}{Hard} & 3D zebrafish cell seg. & Confocal AiryScan & 3D & Static & 588 & $\bullet$ & $\bullet$ & & & & & & & & & \\
       & SARS-CoV-2 / Golgi coloc. & Confocal & 2D & Static & 246 & & $\bullet$ & $\bullet$ & & $\bullet$ & & & & & & \\
       & Ph-cont. bacteria tracking & Phase contrast & 2D & Time-lapse & 482 & $\bullet$ & & & & & $\bullet$ & & & & & \\
       & 4D tile stitching & Confocal ($8\times8$ tiles) & 3D & Static & 1620 & $\bullet$ & $\bullet$ & & & & & & & & $\bullet$ & \\
       & 3D LSFM brain vessel seg. & Light-sheet fluorescence & 3D & Static & 1504 & $\bullet$ & $\bullet$ & & & & & & & & & \\
       & DNA repair foci coloc. & Confocal spinning-disk & 2D & Time-lapse & 733 & & & $\bullet$ & $\bullet$ & $\bullet$ & $\bullet$ & & & & & \\
       & 5D nuclear pore quant. & Confocal & 3D & Time-lapse & 2190 & $\bullet$ & $\bullet$ & $\bullet$ & & & & & & & & \\
       & 3D oncogenic puncta quant. & Confocal spinning-disk & 3D & Static & 5089 & $\bullet$ & $\bullet$ & $\bullet$ & $\bullet$ & & & & & & & \\
\bottomrule
\end{tabular}%
}
\end{sidewaystable}
\begin{sidewaystable}[p]
\centering
\caption{\textbf{Output specifications and scoring metrics per task.} Tasks are grouped by difficulty level and labeled as in Table~\ref{tab:tasks}. \emph{Channels} is the number of imaging channels. \emph{Output} is the required filename pattern. A submission that writes no matching file receives an outcome score of zero (its process score stays visible). Imaging modality, dimensionality, and the subtasks each task exercises are in Table~\ref{tab:tasks}. Composite metrics are written as the weighted sum the task's rubric declares. Metric acronyms: PQ (Panoptic Quality), IoU (Intersection over Union), MAE (Mean Absolute Error), KS (Kolmogorov--Smirnov statistic), and SEG / TRA (the Cell Tracking Challenge segmentation and tracking measures). For stitching, \emph{mosaic} is the rubric's stitching-quality composite (half canvas-shape match, half band intensity-profile correlation against the reference stack) and \emph{feature alignment} is the mean relative error over the six reference (feature, population) pairs.}\label{tab:taskspec}
{\small%
\setlength{\tabcolsep}{1.8pt}%
\renewcommand{\arraystretch}{1.2}%
\begin{tabular}{@{}clcll@{}}
\toprule
Difficulty & Task & Channels & Output & Metric \\
\midrule
\multirow{4}{*}{Easy} & NF-$\kappa$B translocation quant. & 2 & \texttt{*per\_well\_summary*.csv} & Dose--response error \\
       & HeLa nuc./cytopl. seg. & 3 & Instance TIFF / compartment       & Dice \\
       & H\&E nuclear seg. & 3 & \texttt{*.tif} (uint16)           & Dice \\
       & IF cell counting & 1 & Per-image CSV (\texttt{X,Y})      & Count accuracy \\
\midrule
\multirow{4}{*}{Medium} & DNA-PAINT SMLM & 1 & \texttt{*LocalizationList*.txt}   & $0.7\,$Jaccard $+\,0.3\,$intensity $r$ \\
       & Wound-healing kymograph & 1 & \texttt{*\_mf\_dxs/dys.tif} + \texttt{*speedKymograph.tif} & Kymograph / velocity-field composite \\
       & Microglia dynamics & 1 & \texttt{*cell\_labels*.csv}       & Mean of localization F1, classification F1 and trend $r$ \\
       & Microtubule seg. & 1 & \texttt{*mask*.tif} (2D binary)   & Dice \\
\midrule
\multirow{8}{*}{Hard} & 3D zebrafish cell seg. & 1 & \texttt{*\_seg.tif} (+ CSV)       & 3D Dice \\
       & SARS-CoV-2 / Golgi coloc. & 3 & \texttt{*colocalization\_results*.csv} & $0.7\,$significance $+\,0.3\,$direction \\
       & Ph-cont. bacteria tracking & 1 & \texttt{*\_RES.ome.tif} + \texttt{*track*.txt} & $0.5\,$SEG $+\,0.5\,$TRA (CTC) \\
       & 4D tile stitching & 4 & \texttt{*stitch*.tif} + summary CSV & $0.7\,$mosaic $+\,0.3\,$feature alignment \\
       & 3D LSFM brain vessel seg. & 1 & \texttt{*mask*.tif} (3D binary)   & Volumetric Dice \\
       & DNA repair foci coloc. & 2 & CSV (\texttt{start\_normalized})  & KS \\
       & 5D nuclear pore quant. & 2 & Per-cell \texttt{.xlsx}           & $0.5\,$onset MAE $+\,0.5\,$Pearson \\
       & 3D oncogenic puncta quant. & 2 & \texttt{*cell\_quants*.csv}       & Per-nucleus count MAE \\
\bottomrule
\end{tabular}%
}
\end{sidewaystable}

\subsection{Easy tasks}

\paragraph{\texorpdfstring{NF-$\kappa$B translocation quantification (NF-$\kappa$B translocation quant.).}{NF-kappaB translocation quantification (NF-kappaB translocation quant.).}}
BBBC014~\cite{ljosa2012bbbc} is a two-channel widefield fluorescence dose--response
screen acquired on a CellCard reader at $10\times$ magnification ($1360\times1024$\,px,
8-bit). MCF7 and A549 cells were exposed to $12$ concentrations of TNF$\alpha$ with $4$
replicate wells each, a stimulus that drives the transcription factor NF-$\kappa$B (FITC,
in practice a whole-cell stain) from the cytoplasm into the DAPI-counterstained nucleus.
We use the plate in full: $96$ wells $\times$ $2$ channels $=$ $192$ static 2D images,
together with the platemap that links each well to its dose and cell line. The agent is
asked to segment nuclei in the DAPI channel and cell bodies in the FITC channel and to
return one per-well table of the mean nucleus-to-cytoplasm NF-$\kappa$B ratio, which we
score for dose monotonicity, replicate consistency, and separation between untreated and
maximally stimulated wells, separately for each cell line.

\paragraph{HeLa nucleus/cytoplasm segmentation (HeLa nuc./cytopl. seg.).}
HeLaCytoNuc~\cite{helacytonuc} is a fluorescence dataset of HeLa cells (ATCC CCL-2)
fixed and co-stained with DAPI for the nucleus and fluorescent phalloidin for F-actin,
assembled as a technical calibration set for a large-scale high-content RNAi
screen~\cite{ramo2014rnai}. The images are static 2D 8-bit RGB TIFFs
($520\times696$\,px, $0.645$\,$\mu$m pixel size) in which the blue and red channels
carry the nuclear and cytoplasmic stains and the green channel is empty. Of the
$2{,}676$ images in the collection we keep the $265$ ($\approx$$10\%$) that make up the
held-out test subset, the only ones whose nucleus and cytoplasm instance masks were
delineated manually by a specialist instead of being generated with CellProfiler. The
agent must instance-segment the two compartments separately and return, per image, a
nuclear and a cytoplasmic label mask with corresponding instance identities; we score
the mean Dice over the two compartments.

\paragraph{H\&E nuclear segmentation (H\&E nuclear seg.).}
NuInsSeg~\cite{mahbod2024nuinsseg} is a fully annotated dataset for nucleus instance
segmentation in brightfield histology, comprising $665$ H\&E-stained images from $31$
human and mouse organs with manual annotations verified by expert pathologists. We
sample $10$ of them ($\approx$$1.5\%$), five from human kidney and five from human
pancreas: static 2D RGB images of $512\times512$\,px acquired at $20\times$
($0.275$\,$\mu$m pixel size). The two organs were chosen for their histological
contrast, since nuclear size, shape, chromatin staining and packing density range from
isolated epithelial nuclei in tubules and acini to tightly packed lymphocyte
infiltrates. For each image the agent must return a single-channel instance-label mask
covering every nucleus, scored by pixel-wise Dice, with instance-level average precision and panoptic quality reported as diagnostics.

\paragraph{Neural progenitor cell IF counting (IF cell counting).}
CellFMCount~\cite{mohammed2025cellfmcount} is a large-scale benchmark for automated
cell counting: $3{,}023$ single-channel immunocytochemistry fluorescence images of
neural progenitor cells at various proliferation and differentiation stages, with more
than $430{,}000$ manually placed centroid annotations and counts ranging from $0$ to
$2{,}126$ cells per image. We use $21$ static 2D images ($1600\times1200$\,px, 8-bit;
$\approx$$0.7\%$ of the collection) whose reference counts span $0$--$141$ cells, so
that both dense colonies with touching and overlapping cells and genuinely empty fields
are represented. The agent must report the pixel coordinates of every cell centroid in
one CSV per image; predictions are matched to annotations within $10$\,px and scored on
per-image count accuracy, which deliberately penalizes hallucinated detections in the
empty fields.

\subsection{Medium tasks}

\paragraph{DNA-PAINT super-resolution single-molecule localization (DNA-PAINT SMLM).}
A TIRF widefield DNA-PAINT acquisition of a GATTA-PAINT $80$RG DNA-origami nanoruler,
distributed with a hands-on tutorial on computational analysis for single-molecule
localization microscopy~\cite{martens2022smlm}, in which transient hybridization of
imager strands to docking strands makes single fluorophores blink on and off. The input
is one single-channel 16-bit substack of $300$ consecutive 2D frames; although the data
are a movie, the frames encode stochastic blinking rather than a biological time course
and are pooled into a single localization list, so we treat the task as static. The
agent must subtract the fluctuating background, detect the individual blinking events in
every frame and fit their sub-pixel $(x,y)$ positions, returning one table of frame
index, coordinates and integrated intensity. Against a reference list of $1{,}723$
localizations, predictions are matched frame by frame within $1$\,px and scored by a
composite of the Jaccard index ($0.7$) and the intensity correlation ($0.3$).

\paragraph{Wound healing collective migration kymograph (Wound-healing kymograph).}
A live-cell DIC collection of $31$ \emph{in vitro} scratch-wound assays~\cite{zaritsky2015wound} probing the
effect of hepatocyte growth factor/scatter factor (HGF/SF) on collective cell
migration~\cite{zaritsky2012hgf}. We retain one representative experiment per
condition ($6$ of $31$, $\approx$$19\%$), covering MDCK epithelial cells (Control,
$+$HGF/SF) and the DA3 mammary adenocarcinoma line (Control, $+$HGF/SF, $+$PHA,
$+$PHA$+$HGF/SF). Each input is a single-channel 2D time-lapse of $60$--$200$ frames
($1024\times1024$\,px, $0.879$--$1.24$\,$\mu$m/px) acquired every $14.5$\,min, together
with a binary mask of the monolayer at $t=0$ that delimits the wound gap. Following the
authors' reference pipeline, the agent must estimate dense velocity fields between
consecutive frames and assemble, per experiment, a \emph{speed kymograph}: time on the
$x$-axis, distance from the wound leading edge on the $y$-axis, and mean cell speed
($\mu$m/h) as the value. Scoring is dominated by the correlation between the submitted
and reference kymographs. This is the one task whose required output is itself a
visualization, and the only one that exercises feature extraction without a
segmentation or detection subtask.

\paragraph{Phase-contrast microglia activation dynamics (Microglia dynamics).}
BBBC054~\cite{ljosa2012bbbc} follows immortalized mouse microglia (IMG) undergoing
LPS-induced activation by label-free $20\times$ phase-contrast imaging, with expert
annotations assigning each cell at each time point to one of three activation
phenotypes: `round', `amoeboid' or `ramified' (`stratified' in the original
nomenclature). We use one full replicate series, a single-channel 2D time-lapse of $60$
frames ($1280\times1080$\,px, 16-bit) acquired every $30$\,min over $30$\,h, against
which the reference annotation provides ${\sim}58{,}000$ labeled cells. The agent must
detect every cell body and assign its phenotype in each frame, returning one table of
frame index, centroid coordinates and predicted class. Submissions are matched to the
annotation within $15$\,px and scored on the mean of localization F1, classification
macro-F1 and the correlation of the resulting population-shift curve with the reference
time course, so that the score rewards recovering the activation trajectory as well as the cells.

\paragraph{Synthetic microtubule segmentation (Microtubule seg.).}
MicSim\_FluoMT~\cite{bouvrais2025microtubules} is a fully synthetic benchmark released
in support of an attention-based microtubule-segmentation
method~\cite{aitlaydi2025microtubules}. Images are generated in silico, with
Cytosim~\cite{nedelec2007cytosim} providing filament mechanics and
ConfocalGN~\cite{dmitrieff2017confocalgn} simulating confocal fluorescence optics,
including Poisson and Gaussian noise, non-uniform illumination and a severe
foreground/background imbalance (filaments occupy $<5\%$ of the pixels). We draw $16$
single-channel static 2D images ($666\times666$\,px, 8-bit) from the `hard' variant, in
which fluorescence decays toward the filament ends. The agent
must return one binary mask per image covering the thin, curvilinear filaments, which
may extend across the entire field of view as a single connected structure; scoring is
by mean Dice with the topology-aware clDice as tiebreaker.

\subsection{Hard tasks}

\paragraph{3D zebrafish lateral line cell segmentation (3D zebrafish cell seg.).}
Confocal fluorescence stacks of the zebrafish posterior lateral-line primordium, a
migratory tissue that periodically assembles and deposits the rosette-shaped cell
clusters that become mechanosensory organs~\cite{hartmann2020lateralline}. Embryos
(32--36\,hpf) carry the \emph{cldnb:lyn-EGFP} transgene, so the single fluorescent
channel outlines plasma membranes while cell interiors stay dark; stacks were recorded
on an LSM880 in AiryScan FAST mode at anisotropic voxel size
($0.099\times0.099\times0.225$\,$\mu$m), deconvolved and exported as 8-bit TIFFs. We use
three primordia from the study's IDR collection, each a static 3D volume of
$127$--$157$ $z$-slices, in which cell morphology ranges from flat leader cells at the
migratory front to columnar rosette cells at the rear. The agent must return one 3D
instance-label volume per stack, scored against the published reference segmentations by Dice, with instance-level average precision breaking ties; an accompanying per-cell morphometric table is
requested but not scored.

\paragraph{SARS-CoV-2 \& Golgi apparatus colocalization (SARS-CoV-2 / Golgi coloc.).}
Confocal immunofluorescence of fixed Vero cells infected with SARS-CoV-2, released with
a study of viral assembly at the Golgi/ERGIC~\cite{scherer2022sarscov2}. Cells were
immunostained for the viral Spike protein, the viral Nucleocapsid protein and a Golgi
marker, imaged at $70.6$\,nm pixel size and automatically cropped around each single
cell. We use $214$ of these static 2D three-channel crops, distributed over the four
infection-stage conditions as $111$ control, $20$ stage~1, $24$ stage~2 and $59$
stage~3 cells; as infection progresses, newly synthesized Spike accumulates at the
Golgi/ERGIC and its colocalization with the Golgi marker increases. The agent must
compute a per-cell colocalization coefficient for the Spike--Golgi pair and report the
three pairwise comparisons between consecutive conditions in a single table. Scoring
combines whether each comparison reproduces the published significance pattern (control
vs. stage~1 not significant, the later two significant) with whether the per-condition
medians are ordered in the published direction, so that recovering a difference in the
wrong direction does not earn credit.

\paragraph{Phase-contrast microbial single-cell tracking (Ph-cont. bacteria tracking).}
TOIAM (Tracking One-in-a-Million)~\cite{seiffarth2024toiam} is a large-scale benchmark
for microbial live-cell imaging: five $800$-frame phase-contrast time-lapse sequences of
\emph{Corynebacterium glutamicum} growing in microfluidic monolayer cultivation
chambers, densely annotated with more than $1.4$ million segmentation masks, $29$k
tracks and $14$k divisions. We use one of the five sequences ($20\%$), a single-channel
2D time-lapse of $800$ frames ($1094\times938$\,px, 8-bit, $0.072$\,$\mu$m/px) acquired
once per minute. The agent must segment every cell in every frame and link the instances
into lineages, returning a label stack and a Cell Tracking Challenge lineage file whose
track identities match the mask labels. The difficulty is specific to this organism and
to the colony regime: \emph{C. glutamicum} divides by ``snapping'', the colony grows
exponentially so that divisions become increasingly frequent relative to ordinary
frame-to-frame links, and cells start leaving the field of view once the population
exceeds the chamber capacity. We score the task in Cell Tracking Challenge format as
$0.5\,\text{SEG} + 0.5\,\text{TRA}$.

\paragraph{3D multichannel confocal tile stitching (4D tile stitching).}
A confocal laser-scanning dataset of a circulating-tumor-cell model system, in
which white blood cells (WBCs) and the MCF7 breast-cancer line are imaged for
image-cytometry discrimination~\cite{futia2016cytometry}. Four fluorescence channels
label DNA (DAPI), neutral lipids (Bodipy), the epithelial marker Pan-Cytokeratin
(positive in the cancer cells) and the leukocyte antigen CD45 (positive in WBCs). Of the
samples released with the study (WBC only, MCF7 only, and their mixture) we use the
$1{:}1$ WBC:MCF7 mixture, the only one in which both populations must be told apart
within a single field: a static $8\times8$ mosaic of $64$ unstitched tiles, each a
3-slice $z$-stack ($513\times513$\,px per tile, $0.415$\,$\mu$m/px in XY, $5$\,$\mu$m
$z$-spacing, uint16), covering ${\sim}3.1\,\text{mm}\times3.1\,\text{mm}$ and
${\sim}1{,}000$--$2{,}000$ cells. Tile overlap is only a few percent, so registration
must be sub-pixel. The agent must stitch the tiles into a seamless four-channel
composite without collapsing $z$, segment individual cells on the projection, and
extract per-cell image-cytometry features (total signal, spatial second moment
$\langle r\rangle$, spatial-frequency second moment $\langle r_f\rangle$, and their
product $\langle M\rangle$) in each channel. Scoring combines the geometry of the
stitched mosaic with whether the WBC-vs-MCF7 differences follow the direction of the
published reference values~\cite{futia2016cytometry}.

\paragraph{3D LSFM brain capillary segmentation (3D LSFM brain vessel seg.).}
Raw 3D light-sheet fluorescence volumes of cleared, immunolabeled mouse brain released
with the VesselExpress pipeline for quantitative microvasculature
analysis~\cite{spangenberg2023vessels}, where a single fluorescent channel marks the
vascular lumen as bright tubular structures against dark tissue. From the multi-organ
collection we take the striatum of the right hemisphere in the six control animals, i.e.,
six static 3D stacks of $500\times500\times501$ voxels (16-bit), in which vessel
diameters span from a few to tens of voxels. The agent must return one binary 3D vessel
mask per volume and follow it with skeleton-based morphometry (length, branching,
diameter, tortuosity, vessel volume fraction). Masks are compared voxel-wise with the
pipeline's reference segmentations, ranked by Dice with the topology-aware clDice as
tiebreaker; the morphometric table is reported but not yet scored.

\paragraph{DNA repair foci colocalization (DNA repair foci coloc.).}
Live-cell spinning-disk confocal time-lapses of U2OS cells stably co-expressing
TagRFP-PCNA, which marks active replication forks throughout S-phase, and GFP-RAD18,
which is recruited to stalled forks, taken from the primary image data of a study of
53BP1 nuclear bodies~\cite{spies2019foci}. We use the two movies underlying the
published figure, one per condition: mock-depleted control ($157$ frames) and
53BP1-depleted cells ($87$ frames). Each is a 2D maximum-intensity projection time
series at $60\times$ with ${\sim}10$\,min frame intervals, stored as $1000\times1000$\,px
8-bit pseudocolor frames in which the two markers occupy the red and green channels;
a burnt-in timestamp in the corner has to be masked before spot detection. The
biological question is whether 53BP1 sets the timing of RAD18 recruitment, in which case
the colocalization onset normalized by S-phase length should be later in control than in
53BP1-depleted cells. The agent must detect and track the foci, decide when the two
markers first colocalize per cell, and return one table of normalized onset times;
scoring compares the per-condition onset distributions with the reference measurements
($56$ events per condition) through a Kolmogorov--Smirnov statistic.

\paragraph{5D nuclear pore reassembly dynamics quantification (5D nuclear pore quant.).}
Live-cell confocal time-lapses from a quantitative map of nuclear pore
assembly~\cite{otsuka2023npc}, in which cells stably expressing GFP-tagged Nup107, a
Y-complex scaffold nucleoporin, were imaged through mitotic exit on an LSM780
($40\times$ 1.2\,NA water objective), with SiR-Hoechst labeling DNA for nuclear
segmentation and cell-cycle staging. We use five single-cell acquisitions from the
study's Nup107 series, each a 5D (TZCYX) uint16 stack of ${\sim}244$--$293$ time points,
$21$ $z$-slices and $2$ channels, at $0.25$\,$\mu$m lateral and $1$\,$\mu$m axial
sampling with $30$\,s between frames. The agent must segment the two reforming daughter
nuclei in 3D at every time point, detect the frame of anaphase onset, and report per
daughter nucleus the Nup107 recruitment kinetics in the inner-core and non-core regions
of the nuclear envelope together with nuclear volume and surface area. Submissions are
compared with the published per-cell reference tables through a composite of
anaphase-onset error and the correlation of the kinetic curves.

\paragraph{3D oncogenic puncta quantification (3D oncogenic puncta quant.).}
Spinning-disk confocal $z$-stacks of HEK293T cells transiently transfected with
GFP-tagged ZFTA::RELA constructs, the fusion oncoprotein of supratentorial ependymoma,
which forms nuclear condensate-like puncta~\cite{arabzade2025zfta}. Two channels are
acquired, Hoechst for the nuclear counterstain and mEGFP for the fusion protein, at
$100\times$ with $0.2$\,$\mu$m $z$-spacing over $12.2$\,$\mu$m of depth. We use $27$
static 3D fields of view spanning the three constructs whose point mutations in
intrinsically disordered region 3 modulate condensate formation: wild type ($10$
positions), the 3DBD mutant ($8$) and the C129A/H144A double mutant ($9$). The agent
must correct the background, segment nuclei in 3D (a maximum-intensity projection is
insufficient, since nuclear $z$ centroids are required), and quantify the number and
morphology of GFP puncta per nucleus in a single table. Predicted nuclei are paired with
those of the PunctaTools~\cite{baggett2022punctatools} reference analysis within
$5$\,$\mu$m and scored on the per-nucleus error in puncta count, normalized per
condition; unmatched nuclei on either side are charged in full, so neither missing nor
over-segmenting nuclei can improve the score.

\FloatBarrier
\section{Example task instructions}\label{note:2}\label{app:instructions}
Every prompt an agent receives is assembled by one function,
\texttt{render\_instruction} in \texttt{task\_spec.py}, from the task's
\texttt{task\_spec.yaml}: a level-specific \emph{body} (the \texttt{basic} or
\texttt{expert} text), followed by a footer that is identical across levels and
is rendered from the same source for every task. The footer has four blocks:
the required-output contract (file names, formats and label conventions the
evaluator looks for), the source-study policy, a one-line disclosure of the
compute environment, and the absolute input and output paths of the run. No
adapter adds text of its own; each archives the rendered prompt beside the run
(\texttt{<agent>\_instruction.txt}). Across the $16$ tasks the brief bodies are $127$ to $398$ words (median
$191$) and the detailed bodies $233$ to $903$ words (median $461$); the footer
adds $272$ to $530$ words depending on how many deliverables a task declares.

We reproduce both levels for the HeLa nucleus--cytoplasm segmentation task
(Easy level), whose instruction pair is the shortest in the suite; the structure
is the same for every task. The text is copied from the archived prompts of two
DeepSeek Harness runs (brief, \texttt{run\_20260827\_\allowbreak 044030}; detailed,
\texttt{run\_20260831\_\allowbreak 212513}) and is byte-identical to a fresh render, except
that the cluster-specific path prefix is shown as \texttt{<benchmark\_root>}
and the run's own output directory as \texttt{<run\_dir>}; long lines are
wrapped to the text width.

\paragraph{Brief instruction body.}
Written in the voice of a biologist describing the experiment and the
measurement wanted, with no computational guidance.
\begin{lstlisting}[style=instruction]
I'm working with HeLa cells imaged by fluorescence microscopy. I've stained the nuclei with DAPI (blue) and the actin cytoskeleton with phalloidin (red) to visualize the cell body. The images have a pixel size of 0.645 µm.
I need you to identify and outline each individual cell in every image — both the nucleus and the surrounding cytoplasm separately. HeLa cells are an immortalized cervical cancer line, so expect irregular, sometimes elongated morphologies, and cells at different cell cycle stages (which affects nuclear and cytoplasmic size). Cells in culture can be touching or partially overlapping, particularly in denser areas of the well. Please provide visualizations to help me understand and verify the results (e.g. representative examples of the input data and processed output; overlays or side-by-side comparisons; plots with legends; ...). Publication-ready visualizations are also highly appreciated.
\end{lstlisting}

\paragraph{Detailed instruction body.}
Adds a recommended pipeline, channel assignments, candidate tools and the
annotation conventions the reference follows, while leaving every
implementation choice to the agent.
\begin{lstlisting}[style=instruction]
Act as a Bioimage Analyst. Perform dual-compartment instance segmentation (nuclei + cytoplasm) on 8-bit RGB fluorescence images of HeLa cells.
1. Channel Assignment: Images are 3-channel RGB TIFs (520x696 px, 8-bit) with a pixel size of 0.645 µm.
   - Red channel: phalloidin-stained F-actin — use for cytoplasm segmentation.
   - Blue channel: DAPI-stained nuclei — use for nuclear segmentation.
   - Green channel: unused — ignore.
   Extract individual channels before processing (do not operate on the merged RGB).

2. Nuclear Segmentation:
   - Apply illumination correction (e.g., rolling-ball background subtraction)
     on the blue channel to compensate for field-of-view intensity gradients.
   - Threshold with Otsu or a local adaptive method; apply binary fill-holes and
     morphological opening to remove debris.
   - Separate touching nuclei via distance-transform watershed or a pretrained
     model (e.g., StarDist, Cellpose with nucleus weights).
   - Label each nucleus with a unique positive integer; background = 0.

3. Cytoplasm Segmentation:
   - Use the nuclear instances as seeds for a marker-controlled watershed on the
     red (actin) channel. This enforces nucleus-cytoplasm correspondence and
     prevents label mismatch.
   - Alternatively, run Cellpose (cyto2/cyto3 model) on the red channel and
     match resulting instances to nuclear labels by maximum IoU overlap.
   - The cytoplasm region should represent the full cell body minus the nucleus
     (i.e., the annular cytoplasmic ring), or the full cell body depending on
     the downstream use — here, include the nucleus within the cytoplasm mask,
     following the CellProfiler convention of the source study.
   - Enforce that cytoplasm instance ID i corresponds to nucleus instance ID i
     for the same cell.

4. Border Cells: Include cells touching image borders consistently; do not
   discard them — the source study's annotation convention retains
   border-touching instances.

5. Output: produce the per-image nuclei and cytoplasm instance-label
   masks as specified in the `deliverables` section below (see there
   for filename patterns, formats, and label conventions). Cytoplasm
   instance id N must correspond to nucleus instance id N for the
   same cell.

6. Quality Control and Visualization:
   - Save representative composite images showing: (a) merged RGB input,
     (b) nuclear segmentation masks with colored label boundaries overlaid on
     the blue (DAPI) channel, (c) cytoplasm segmentation masks with colored
     boundaries overlaid on the red (actin) channel.
   - Generate a scatter plot showing nuclear area vs. cytoplasmic area for all
     segmented cells to assess the nucleus-to-cytoplasm ratio distribution.
   - Include scale bars and clear legends identifying segmentation classes.
\end{lstlisting}

\paragraph{Shared footer.}
Appended unchanged to both bodies. Its second block is the source-study policy:
the agent may consult methods literature and software documentation but may not
retrieve the source publication or its reported values, nor identify it from
file names or metadata.
\begin{lstlisting}[style=instruction]
---
Required output files (write these inside the output directory; the evaluator will look for them by name):

- nuclei_masks  (required, multi-file (one per input sample / sequence))
  Filename pattern: *nuclei*.tif*
  Format: uint8 TIF (instance labels)
  Per-image nuclear instance-label masks. Each filename must contain the substring 'nuclei' (e.g. 'nuclei_masks/N.tif' or 'N_nuclei.tif') so the evaluator can disambiguate from cytoplasm masks. Stems must match the input image stems. uint8, integer labels, 0 = background.

- cytoplasm_masks  (required, multi-file (one per input sample / sequence))
  Filename pattern: *cytoplasm*.tif*
  Format: uint8 TIF (instance labels)
  Per-image cytoplasm instance-label masks. Each filename must contain 'cytoplasm' (e.g. 'cytoplasm_masks/N.tif' or 'N_cytoplasm.tif'). uint8, integer labels, 0 = background. Cytoplasm instance id N must correspond to nucleus instance id N for the same cell.


---
Source-study policy:
- This task is derived from a published study. Do not search for, retrieve, or read that source publication, its figures, supplementary materials, or its reported values, and do not try to identify it from file names or metadata. Derive every reported number from the provided data itself; copying values from the source study invalidates the analysis. General background knowledge, methods literature, and software documentation (e.g. library or tool docs) may be used freely.

---
Compute environment:
- This machine has a CUDA-capable GPU available. Prefer GPU-accelerated execution for compute-heavy steps such as deep-learning model inference -- it is typically far faster than CPU. If a particular library or model does not support the GPU, fall back to CPU.

---
I/O paths (set by the benchmark runner):
- Input directory (read-only): <benchmark_root>/benchmark_tasks/fluo-helacytonuc-cell-segmentation/input
  All input data for this task lives under this directory. Discover the layout yourself (e.g. with `Path.iterdir` / `rglob`); the benchmark intentionally does not enumerate files for you.
- Output directory (write here): <run_dir>
  Save every final deliverable inside this directory. Files written anywhere else are invisible to the evaluator.
Use absolute paths when reading inputs and writing outputs.
\end{lstlisting}

\FloatBarrier
\section{Agent integration and run limits}\label{note:3}\label{app:adapters}
\textbf{Biomni} (LLM tool-use) writes and executes Python in a conda environment;
provider token usage is captured at the model-invocation layer. \textbf{Claude
Code} and \textbf{Codex}, both command-line interfaces (CLIs), are driven through their JSON event streams, from
which native token usage is parsed (input, output and cache-read tokens kept
separate so that cached reads are not double-counted); Claude Code reaches the
study's models through its base-URL setting, Codex through its provider
configuration and \textbf{DeepSeek Harness} natively. \textbf{CopilotJ} drives
Fiji/ImageJ through a GUI bridge under a virtual display. \textbf{Agentic-J}, a
second Fiji/ImageJ GUI agent, runs inside a containerized desktop (Docker or
Apptainer) with an in-session LLM through the same contract.

The only agent-specific inner limit is Biomni's per-step execution timeout, set
so that a heavy step such as a deep segmenter over a whole image set completes
while a hung step cannot consume the whole wall-clock budget. The coding CLIs
and the two GUI agents impose no uniform turn cap; turn and tool-call counts
are logged as diagnostics.

The external wall-clock budget was $4$\,h. Most model-exchange sessions (Kimi
K2.6, GLM-5.1 and Claude Opus~5) ran with $2$\,h, so their timed-out runs stop
at $120$\,min.

\FloatBarrier
\section{Baseline and reference submissions}\label{note:4}\label{app:gtreplay}
To confirm that a perfect submission scores $1.0$, each task's own ground truth
is fed back through the evaluation as if it were the agent's prediction. It
reaches $\approx 1.0$ on every task with a ground-truth file. No metric is
therefore capped below $1$, and the declared output format admits a perfect
submission.

Most tasks are satisfied by the verbatim ground-truth files. Five tasks require an output derived from the ground truth (the brain-microvessel masks, the HeLa nucleus and cytoplasm compartments, and the microglia, DNA-repair-foci and SARS-CoV-2-Golgi tables), and a per-task builder emits the format-correct perfect submission for them. The NF-$\kappa$B translocation task cannot be replayed because it has no ground-truth file. Instead, its outcome score is computed from the dose--response properties of the submitted table (a monotone
rise of the nuclear-to-cytoplasmic ratio with dose, agreement between replicate
wells and a significant difference from the untreated control), each normalized to $[0,1]$, so this metric can still reach $1$.

Two synthetic submissions are scored per task with the same pipeline that scores
agents, with no agent and no GPU involved. The first is an empty output
directory. Its outcome score is zero on every task, and the judge decides no
rubric item on twelve tasks and at most eight (all but one failed) on the other
four, so it fixes the zero point. The second is a mock submission containing only a report that
describes a complete pipeline, with no output file. On every task the judge
finds no evidence for any rubric item and decides none, so the submission
receives no process score, and its outcome score is zero. The pair bounds
what a submission with no analysis earns on each score.

\FloatBarrier
\section{Process rubric and judge validation}\label{note:5}\label{app:judge}
The process score is a severity-weighted rubric over the pipeline sections
(load, preprocess, segment or detect, measure, statistics, visualize, report),
applied by a vision--language model to the rendered outputs of a submission.
The judge was checked against a human expert on two questions: whether it agrees
with the expert and whether it systematically over-credits.

\paragraph{The rubric.}
The rubric is a single file of $105$ yes/no items, reproduced in full in
Table~\ref{tab:rubric}. Fifty items are general. Forty-five apply to every run (input understanding $9$, tool choice and use $9$, reporting to the user $6$, quantification $7$, basic visualization $14$), and five advanced-visualization items apply to the eight tasks whose deliverable is quantitative.
The remaining $55$ items sit in ten task-specific subsections (stitching,
segmentation, feature extraction, classification, statistical plotting,
filament extraction, denoising, spot detection, tracking, colocalization);
each task's own rubric file names the subsections it exercises, and every item
in a named subsection is scored. A run is therefore judged on between $49$ and
$75$ items. Each item carries a severity that sets its weight (critical $3$,
major $2$, minor $1$); the process score is the weighted fraction of passed
items among those the judge decided (Section~\ref{sec:process}).

\paragraph{Judge prompt and evidence.}
The judge is called once per pair of rubric items, with a system prompt that
is identical for every task and a user turn in two parts: a submission-wide
context block, byte-identical across all calls for a submission and served
from the provider's prompt cache, and a short chunk listing
the two items to decide. The system prompt is reproduced verbatim below; the
user turn is shown as a skeleton with placeholders in angle brackets, filled
here with the first two segmentation items of the HeLa task. The evidence the
judge sees is assembled from the run folder by file type. The agent's scripts
(any of Python, Groovy, ImageJ macro, Java, R, MATLAB or shell), Markdown
reports and the harness transcript are always attached, capped at $12{,}000$, $8{,}000$ and $12{,}000$ characters per file
within a $60{,}000$-character total; other text files are searched by keyword
and only the matching lines are attached to the item that matched. Files the
harness writes about itself (run manifests, step logs, the completion
sentinel) are excluded.
Up to eight of the agent's PNG or JPEG figures are attached, chosen after
dropping the files that match the task's deliverable pattern (those are scored
deterministically) and collapsing per-image batches to two representatives.
Reference
images from the task's evaluation folder are attached only when a directly
viewable one exists (two tasks); the reference label stacks are never shown.
One task (Golgi colocalization) appends four task-specific anchor rules to the
system prompt, and one evaluator (3D puncta) exports its computed metrics
into the context block; both are listed in the repository.

The judge returns one JSON decision per item with a status, a confidence, a rationale and the
evidence files it relied on; a pass that cites no evidence file is kept but
flagged, and a decision of \emph{unknown} must name one of four reasons (no
relevant evidence, ambiguous evidence, rubric unclear, image unreadable). The
judge was run with a single sample per chunk at temperature $0.1$; decisions
are cached by a hash of the prompt and the attached images, and re-scoring a
run reuses the cache.

\begin{lstlisting}[style=instruction]
# System prompt (verbatim; Checklist.yaml and the evaluator source are the reference)
You are a bioimage analysis benchmark judge. For each checklist item you must decide pass / fail / unknown based on the evidence provided (agent-produced code/reports/CSVs and agent image outputs, optionally accompanied by ground-truth reference images).

Decision rules:
- 'pass': concrete evidence in the attached text snippets OR images shows the action was performed correctly. Strongly prefer to cite at least one evidence_ref; missing refs do not invalidate a pass but will reduce its confidence weighting downstream.
- 'fail': evidence clearly contradicts the action (e.g. wrong channels, wrong metric, missing step).
- 'unknown': ONLY when the evidence genuinely does not support either a pass or a fail. Do NOT use 'unknown' as a 'safer' default; if you can see the expected artifact in an image or snippet, pick pass/fail. Always supply unknown_reason from the allowed enum.

Allowed unknown_reason values: ['no_relevant_evidence', 'ambiguous_evidence', 'rubric_unclear', 'image_unreadable']
Return STRICT JSON only (no markdown, no prose outside JSON) with shape:
{"results":[{"item_id":..., "vlm_status":..., "vlm_confidence":..., "vlm_rationale":..., "vlm_evidence_refs":[...], "unknown_reason":...}]}

# Appended only for tasks whose rubric file declares vlm_anchors (one task in the suite)

TASK-SPECIFIC ANCHOR RULES (override generic guidance above for this task):
- <rule 1>
- <rule 2>
...
\end{lstlisting}
\begin{lstlisting}[style=instruction]
# User turn, part 1: shared context (identical for every chunk of a submission,
# served from the provider's prompt cache after the first chunk)

TASK INSTRUCTION (for context only):
<the rendered instruction the agent received>

IMAGE ORDER (attached below, in order):
  [IMG 1] [GROUND TRUTH] <reference image, only if a directly viewable one exists>
  [IMG 2] [AGENT OUTPUT] <selected figure 1>
  [IMG 3] [AGENT OUTPUT] <selected figure 2>
  ...                       (at most 8 agent images and 4 reference images)

AGENT ARTIFACTS (shared evidence for every checklist item below):
  --- result_metrics/evaluator_evidence.txt ---
  STRUCTURED RESULT METRICS (computed by the Python evaluator, use as ground
  truth for any item that depends on these values):
  - <key>: <value>                (only for tasks whose evaluator exports them)
  --- agent_report/<file>.md ---
  <report text, first 8,000 characters>
  --- agent_code/<file>.py ---
  <script text, first 12,000 characters per file>
  --- agent_log/<agent>_log.txt ---
  <transcript, de-noised, 12,000 characters (head and tail)>

# User turn, part 2: the chunk (two items per request)

CHECKLIST ITEMS TO JUDGE (one JSON decision per item_id):
- item_id: chk_0011_did_the_agent_choose_an_algorithm_capable_of_separating_touching_objects_e_g_wat
  text:    Did the agent choose an algorithm capable of separating touching objects (e.g., Watershed, StarDist, Cellpose) rather than simple global thresholding?
  section: task_specific
  subsection: segmentation
  evidence_snippets:
    --- supporting/<file> ---
    <keyword-retrieved lines from supporting files not already shown above>

- item_id: chk_0012_did_the_agent_filter_out_objects_that_are_clearly_noise_based_on_size_or_shape
  text:    Did the agent filter out objects that are clearly noise based on size or shape?
  section: task_specific
  subsection: segmentation
  evidence_snippets: (see shared agent artifacts above)

Respond with ONLY the JSON object described in the system prompt. Each item_id must appear exactly once in 'results'.
\end{lstlisting}

\paragraph{Expert review design.}
Twenty runs from the GPT-5.6~Sol study were drawn to cover all six agents
and all $16$ tasks, with four tasks (3D puncta, 5D nuclear-pore kinetics,
mosaic stitching, microglia progression) represented twice. For each run a
blinded copy of the run folder was prepared in which every judge decision was
erased back to \emph{undecided}; the raw input images were omitted for size.
An expert cell biologist among the authors, who had not seen any
judge decision, answered every rubric item
for each run in a review tool that shows the folder's outputs, figures, code
and logs beside the question, with three answers: \emph{yes} (demonstrably
satisfied by an artifact), \emph{no} (demonstrably not, including ``claimed
but no artifact'') and \emph{skip} (undecidable from the folder). The expert
also applied \emph{skip} to items that did not apply to the run, treating each
question as carrying an implicit ``when applicable''. Agreement is computed on
the items both the expert and a judge decided; abstentions on either side are
excluded from agreement and reported as rates. Three candidate judges
were run on the same $20$ submissions with identical prompts, image selection
and settings: Claude Sonnet~5, Claude Opus~5 and Gemini~3.1~Pro. Intervals are
$95\%$ cluster-bootstrap intervals that resample whole runs, because items
within a run are not independent.

\paragraph{Agreement with the expert.}
The expert decided $1{,}085$ of $1{,}273$ items ($767$ yes, $318$ no) and
skipped $188$ ($15\%$). On the $922$ items that both decided, the adopted judge
(Sonnet~5) agreed with the expert on $87\%$ (Cohen's $\kappa=0.67$, $95\%$ CI
$0.61$--$0.73$; Table~\ref{tab:judge}). Its errors were
asymmetric, as it passed $68$ of the $251$ items the expert marked as failed ($27\%$) and
failed $48$ of the $671$ items the expert marked as passed ($7\%$). The two other
candidates were statistically indistinguishable in accuracy and $\kappa$
($0.87$ and $0.66$ for Opus~5; $0.86$ and $0.67$ for Gemini~3.1~Pro) but differed
in bias direction, with Opus~5 the most lenient (over-credit $33\%$) and
Gemini~3.1~Pro the most severe
(under-credit $10\%$). Sonnet~5 was adopted on the combined basis of
agreement with the expert and cost. The judge abstained on $20\%$ of items, the expert
on $15\%$; the judge also abstained on half of the items the expert skipped,
while the expert was able to decide $63\%$ of the items on which the judge
abstained ($96$ yes, $67$ no), consistent with the reviewer opening files that
the judge's evidence selection did not include.

Agreement varied by rubric stratum. It was highest for input understanding
($\kappa=0.87$), statistical plotting ($0.75$) and the task-specific items
($0.72$), and lowest for tool choice and use ($0.47$), where the judge passed
half of the items the expert marked as failed; these are questions such as whether
parameters were chosen by a quantitative criterion or read from the image
instead of being hard-coded, which the judge tends to credit on the strength of the
narration. By severity, agreement on critical items was $97\%$ but $\kappa$ was
only $0.52$ with a wide interval, because only $8$ of $221$ critical items
were failed under expert review and the judge caught four of them.

\paragraph{Run-level process scores.}
Recomputing each run's process score from the expert's labels with the
rubric's own severity weights (decided items only, as in production) gave
scores that correlated with the judge's at $r=0.54$ (Spearman $\rho=0.40$; $r=0.66$
when both are restricted to the items both decided), with the judge higher
by $0.02$ on average. The expert's process score did not predict the outcome
score ($r=-0.03$, $95\%$ CI $-0.48$ to $0.31$; $\rho=-0.27$), whereas the
judge's process score on the same $20$ runs gave $r=0.28$. The five runs with
an outcome below $0.1$ received expert process scores between $0.74$ and
$0.82$. The weak process--outcome relation reported in Section~\ref{sec:res-process}
therefore persists when an expert applies the same rubric.

\paragraph{Rubric items with low discrimination.}
Runs failed $4\%$ of critical items, $31\%$ of major items and $44\%$ of
minor items under expert review (Table~\ref{tab:faileditems}). Critical items carry
$38\%$ of the decided weight but accounted for $6\%$ of the weighted failures;
minor items carry $20\%$ of the weight and accounted for $37\%$ of the failures. The items
failed most often were reporting conveniences (a gallery comparing parameter
settings, $20/20$; a draft figure legend, $17/20$; a color bar, $15/18$) and
two measurement safeguards that almost no run applied: excluding saturated
pixels from intensity measurements ($8/8$) and checking for clipped pixels
before quantifying ($16/17$). Re-weighting did not recover outcome
information. Promoting the four most-failed major methodology items to
critical gave $r=-0.05$ with the outcome, scoring critical items alone gave
$r=0.08$, and scoring the task-specific items alone gave $r=0.02$. The rubric
measures whether an analysis was conducted and reported in a professional
manner, which nearly every run was. Thirteen of the $105$ distinct items
were skipped by the expert in at least half of the runs they appeared in
(for example ``did the agent respect user-specified tool constraints'' when no
constraint was given), and these are candidates for per-task filtering in a
future rubric revision.

\paragraph{Evidence preserved by each harness.}
The expert's skip rate, a measure of how much of the run a reviewer can
reconstruct from the folder, ranged from $8\%$ for Agentic-J and $11$--$13\%$
for Codex, Claude Code and Biomni to $19\%$ for CopilotJ and $20\%$ for the
DeepSeek Harness; the judge's abstention rate follows the same order
($11\%$ to $30\%$). Across all $282$ GPT-5.6~Sol runs, an agent-written script
was preserved in the run folder in $100\%$ of Biomni runs, $96\%$ of Agentic-J
runs, $85\%$ of DeepSeek Harness runs, $69\%$ of Codex runs, $60\%$ of Claude
Code runs and $6\%$ of CopilotJ runs, which drives Fiji through its GUI
and leaves macros rather than scripts. The decided-only aggregation of the
process score (Section~\ref{sec:process}) limits the effect of this variation on the process score. The archived expert labels,
the three judges' decisions and the scripts that produce every number in this
appendix are included in the repository.

\newlength{\rubricw}
{\small\setlength{\tabcolsep}{4pt}%
\setlength{\rubricw}{\dimexpr\linewidth-4\tabcolsep\relax}%
\settowidth{\dimen0}{105}\addtolength{\rubricw}{-\dimen0}%
\settowidth{\dimen0}{Severity}\addtolength{\rubricw}{-\dimen0}%
\renewcommand{\arraystretch}{1.08}
\begin{longtable}{@{}r>{\raggedright\arraybackslash}p{\rubricw}l@{}}
\caption{\textbf{The process rubric.} All $105$ yes/no items, grouped by rubric subsection, with the severity that sets each item's weight (critical $3$, major $2$, minor $1$). Input understanding, tool choice and use, reporting, quantification and basic visualization apply to every task; advanced visualization applies to tasks whose deliverable is quantitative; each task-specific subsection applies to the tasks named under its heading, selected by the task's own rubric file. A run is scored on the items its task selects (between $49$ and $75$ items), as the severity-weighted fraction of passed items among those the judge decided.}\label{tab:rubric}\\
\toprule
\# & Item & Severity \\
\midrule
\endfirsthead
\multicolumn{3}{@{}l}{\footnotesize\emph{Table~\ref{tab:rubric} (continued)}}\\[2pt]
\toprule
\# & Item & Severity \\
\midrule
\endhead
\midrule
\multicolumn{3}{r@{}}{\footnotesize\emph{continued on next page}}\\
\endfoot
\bottomrule
\endlastfoot
\addlinespace[3pt]
\multicolumn{3}{@{}p{\linewidth}@{}}{\textbf{Input understanding} \hspace{0.6em}{\emph{applies to:} all tasks}}\\
\addlinespace[1pt]
1 & Did the agent correctly identify image metadata (2D vs 3D, channels, timepoints)? & critical \\
2 & Did the agent select an appropriate strategy for the request? & critical \\
3 & Did the agent ignore irrelevant image channels? & major \\
4 & Did the agent separate channels correctly? & critical \\
5 & Was the image read with full bit-depth available? & critical \\
6 & Did the agent correctly assign biological meaning to channels? & critical \\
7 & Did the agent quantify on raw data, not contrast-enhanced visualizations? & critical \\
8 & Did the agent ensure raw data was never overwritten? & critical \\
9 & Did the agent check for clipped/saturated pixels before quantification? & major \\
\addlinespace[3pt]
\multicolumn{3}{@{}p{\linewidth}@{}}{\textbf{Tool choice and use} \hspace{0.6em}{\emph{applies to:} all tasks}}\\
\addlinespace[1pt]
10 & Did the agent normalize intensity range when required by a specific tool? & major \\
11 & Did the agent provide reasoning for each chosen tool? & minor \\
12 & Did the agent estimate parameters from visual features rather than using hard-coded defaults? & major \\
13 & Did the agent validate the strategy on a small crop or single slice first? & minor \\
14 & Did the agent perform a parameter sweep before finalizing? & minor \\
15 & Did the agent use a quantitative metric to pick the best parameters? & major \\
16 & Did the agent handle per-image failures gracefully (continue batch, report failure)? & major \\
17 & Did the agent prefer computationally cheap methods when sufficient? & minor \\
18 & Did the agent maintain original bit-depth throughout the pipeline? & critical \\
\addlinespace[3pt]
\multicolumn{3}{@{}p{\linewidth}@{}}{\textbf{Reporting to the user} \hspace{0.6em}{\emph{applies to:} all tasks}}\\
\addlinespace[1pt]
19 & Did the agent provide a gallery/montage comparing different settings? & minor \\
20 & Did the agent respect user-specified tool constraints? & major \\
21 & Did the agent explain errors in plain English? & minor \\
22 & Did the agent warn about potential biases in measurements? & minor \\
23 & Is the generated code documented well enough to serve as a tutorial? & minor \\
24 & Did the agent suggest better alternatives for future runs? & minor \\
\addlinespace[3pt]
\multicolumn{3}{@{}p{\linewidth}@{}}{\textbf{Quantification} \hspace{0.6em}{\emph{applies to:} all tasks}}\\
\addlinespace[1pt]
25 & Are measurement units correct (microns if metadata available)? & major \\
26 & Is the data structure appropriate (DataFrame/CSV, not printed numbers)? & major \\
27 & Do image labels match data table labels? & major \\
28 & Did the agent calculate the specific metrics requested? & critical \\
29 & Did the agent group results correctly (by filename/folder/condition)? & major \\
30 & Did the agent suggest follow-up analyses based on findings? & minor \\
31 & Did the agent justify outlier removal mathematically or biologically? & minor \\
\addlinespace[3pt]
\multicolumn{3}{@{}p{\linewidth}@{}}{\textbf{Visualization (all tasks)} \hspace{0.6em}{\emph{applies to:} all tasks}}\\
\addlinespace[1pt]
32 & Did the agent provide a visualization reference (e.g., RGB overlay of raw/processed data)? & major \\
33 & Did the agent arrange results into a multi-panel figure suitable for a manuscript? & minor \\
34 & Did the agent provide a draft figure legend for each visualizations? & minor \\
35 & Do output filenames relate back to input filenames? & minor \\
36 & Did the agent save images in a scientifically valid format (TIFF, not JPEG)? & major \\
37 & Did the agent generate a Methods description? & minor \\
38 & Did the agent disclose any non-linear adjustments during the figure making process (e.g., Gamma)? & major \\
39 & Did the agent avoid cleaning that could hide artifacts or biological features? & critical \\
40 & Are axis tick labels formatted correctly? & minor \\
41 & Did the agent save figures in vector format (PDF/SVG) or high-DPI raster (300 DPI TIFF)? & minor \\
42 & Is a scale bar present on the final visualization? & major \\
43 & Is the scale bar correct? & major \\
44 & Is the color map appropriate and contrast adjusted? & minor \\
45 & Is the color map represented (color bar)? & minor \\
\addlinespace[3pt]
\multicolumn{3}{@{}p{\linewidth}@{}}{\textbf{Visualization (quantitative tasks only)} \hspace{0.6em}{\emph{applies to:} 3D oncogenic puncta quant., 3D zebrafish cell seg., 3D LSFM brain vessel seg., 5D nuclear pore quant., 4D tile stitching, SARS-CoV-2 / Golgi coloc., DNA repair foci coloc., Microglia dynamics}}\\
\addlinespace[1pt]
46 & Did the agent clearly report N (cells, images, experiments)? & major \\
47 & Did the agent provide a quantitative plot (e.g., box-plot, histogram, scatterplot)? & major \\
48 & Are axes labeled on plots/histograms? & major \\
49 & Is the data clearly marked with a legend? & major \\
50 & Is the plot type correct for the data? & major \\
\addlinespace[3pt]
\multicolumn{3}{@{}p{\linewidth}@{}}{\textbf{Stitching} \hspace{0.6em}{\emph{applies to:} 4D tile stitching}}\\
\addlinespace[1pt]
51 & Did the agent preserve the original bit-depth and pixel size metadata in the stitched output? & critical \\
52 & Did the agent correctly handle z-stacks or multi-dimensional tiles (e.g., projecting or selecting a single z-plane before stitching)? & critical \\
53 & Did the agent save the stitched result in a lossless format (e.g., TIFF) rather than a lossy format (e.g., JPEG)? & critical \\
54 & Did the agent correctly parse tile layout and positions from the file metadata rather than assuming a fixed grid? & minor \\
55 & Did the agent use a registration-based method (phase correlation, feature matching) to compute sub-pixel tile offsets rather than relying solely on stage coordinates? & major \\
56 & Did the agent apply global optimization (e.g., minimum spanning tree, least-squares) to distribute alignment errors rather than chaining pairwise registrations? & minor \\
57 & Are bands or seams visually noticeable at the tile overlaps? & major \\
58 & Did the agent blend overlapping tile regions (linear, multi-band, or feathering) to avoid visible seams? & major \\
59 & Did the agent use a refence channel (e.g. DAPI) for registration and apply the same transforms to all channels? & major \\
60 & Did the agent handle edge tiles or incomplete grids gracefully (no crashes, clear warnings)? & minor \\
\addlinespace[3pt]
\multicolumn{3}{@{}p{\linewidth}@{}}{\textbf{Segmentation} \hspace{0.6em}{\emph{applies to:} 3D oncogenic puncta quant., 3D zebrafish cell seg., 3D LSFM brain vessel seg., 5D nuclear pore quant., 4D tile stitching, NF-$\kappa$B translocation quant., HeLa nuc./cytopl. seg., H\&E nuclear seg., Microglia dynamics, Ph-cont. bacteria tracking, Microtubule seg.}}\\
\addlinespace[1pt]
61 & Did the agent choose an algorithm capable of separating touching objects (e.g., Watershed, StarDist, Cellpose) rather than simple global thresholding? & critical \\
62 & Did the agent filter out objects that are clearly noise based on size or shape? & major \\
63 & Did the agent handle objects touching the image border correctly? & major \\
64 & Did the agent fill holes inside objects if biologically appropriate? & minor \\
65 & Did the agent use distinct integer labels for each object (1, 2, 3...) rather than a binary mask (0, 1)? & critical \\
\addlinespace[3pt]
\multicolumn{3}{@{}p{\linewidth}@{}}{\textbf{Feature extraction} \hspace{0.6em}{\emph{applies to:} 3D oncogenic puncta quant., 3D zebrafish cell seg., 3D LSFM brain vessel seg., 5D nuclear pore quant., 4D tile stitching, NF-$\kappa$B translocation quant., SARS-CoV-2 / Golgi coloc., Microglia dynamics, Wound-healing kymograph}}\\
\addlinespace[1pt]
66 & Did the agent perform background subtraction before measuring intensity? & major \\
67 & Did the agent measure intensity on the raw data, not on a visualization/LUT-adjusted image? & critical \\
68 & Did the agent exclude saturated pixels from mean intensity calculations? & major \\
69 & If measuring shape (e.g., Circularity), did the agent ensure pixels are square or correct for aspect ratio? & major \\
\addlinespace[3pt]
\multicolumn{3}{@{}p{\linewidth}@{}}{\textbf{Classification} \hspace{0.6em}{\emph{applies to:} Microglia dynamics}}\\
\addlinespace[1pt]
70 & Did the agent define clear, biologically motivated class boundaries? & major \\
71 & Did the agent report per-class metrics (precision, recall, F1) rather than only overall accuracy? & major \\
72 & Did the agent use extracted features (not raw pixels) as input to the classifier? & major \\
\addlinespace[3pt]
\multicolumn{3}{@{}p{\linewidth}@{}}{\textbf{Statistical plotting} \hspace{0.6em}{\emph{applies to:} 3D oncogenic puncta quant., 5D nuclear pore quant., NF-$\kappa$B translocation quant., SARS-CoV-2 / Golgi coloc., DNA repair foci coloc., Microglia dynamics}}\\
\addlinespace[1pt]
73 & Is a statistical test performed? & major \\
74 & Is the statistical test appropriate? & critical \\
75 & Did the agent apply identical processing settings to all compared conditions? & critical \\
76 & Did the agent check for normality before using parametric tests? & major \\
77 & Did the agent apply multiple-comparison corrections (e.g., Bonferroni, FDR) when testing multiple hypotheses? & major \\
78 & Did the agent compare control vs. treated groups when filenames/folders suggest a comparative study? & major \\
79 & Did the agent provide a p-value? & major \\
80 & Did the agent provide a distribution plot rather than just an average? & major \\
81 & Did the agent report effect sizes? & minor \\
82 & Is significance annotated in the plot? & minor \\
\addlinespace[3pt]
\multicolumn{3}{@{}p{\linewidth}@{}}{\textbf{Filament extraction} \hspace{0.6em}{\emph{applies to:} 3D LSFM brain vessel seg., Microtubule seg.}}\\
\addlinespace[1pt]
83 & Did the agent perform skeletonization (reducing structures to 1-pixel-wide lines)? & critical \\
84 & Did the agent analyze branching points (nodes) and endpoints? & major \\
85 & Did the agent prune small branches that are likely noise? & minor \\
86 & Did the agent prioritize topological connectivity (avoiding fragmentation of single filaments)? & major \\
\addlinespace[3pt]
\multicolumn{3}{@{}p{\linewidth}@{}}{\textbf{Denoising} \hspace{0.6em}{\emph{applies to:} Microtubule seg.}}\\
\addlinespace[1pt]
87 & Did the agent avoid hallucinating high-frequency details beyond the resolution limit? & critical \\
88 & Did the agent preserve total intensity flux before and after denoising? & major \\
89 & Does the image avoid looking waxy or over-smoothed? & major \\
90 & Are edges sharpened without introducing ringing artifacts? & minor \\
\addlinespace[3pt]
\multicolumn{3}{@{}p{\linewidth}@{}}{\textbf{Spot detection} \hspace{0.6em}{\emph{applies to:} 3D oncogenic puncta quant., IF cell counting, DNA repair foci coloc., DNA-PAINT SMLM}}\\
\addlinespace[1pt]
91 & Did the agent apply a DoG or LoG filter to enhance spots before detection? & major \\
92 & Did the agent perform a local maxima search rather than a global threshold? & major \\
93 & Did the agent provide (X, Y, Z) coordinates for detected spots? & major \\
94 & Did the agent account for Z-spread to avoid double-counting spots across adjacent slices? & major \\
95 & Are there obvious bright spots that were missed (False Negatives)? & major \\
96 & Are there background noise speckles marked as spots (False Positives)? & major \\
\addlinespace[3pt]
\multicolumn{3}{@{}p{\linewidth}@{}}{\textbf{Tracking} \hspace{0.6em}{\emph{applies to:} 5D nuclear pore quant., DNA repair foci coloc., Ph-cont. bacteria tracking}}\\
\addlinespace[1pt]
97 & Did the agent generate a track ID that persists across frames? & critical \\
98 & Did the agent account for cell/object division? & major \\
99 & Did the agent account for cell/object fusion? & major \\
100 & Did the agent handle detection gaps (resuming tracks after missed frames)? & major \\
101 & Did the agent calculate velocity or displacement metrics from the tracks? & minor \\
\addlinespace[3pt]
\multicolumn{3}{@{}p{\linewidth}@{}}{\textbf{Colocalization} \hspace{0.6em}{\emph{applies to:} SARS-CoV-2 / Golgi coloc., DNA repair foci coloc.}}\\
\addlinespace[1pt]
102 & Did the agent analyze colocalization within a specific ROI (e.g., inside the cell) rather than the whole image? & major \\
103 & Did the agent perform a statistical control (e.g., Costes' randomization)? & major \\
104 & Did the agent avoid using Pearson's Correlation on thresholded (binary) data? & critical \\
105 & Did the agent output a scatterplot of Channel 1 vs. Channel 2 intensities? & minor \\
\end{longtable}}

\begin{table}[H]\centering
\caption{\textbf{Agreement between candidate judges and a human expert.} Twenty runs (six harnesses, all $16$ tasks) reviewed item by item by one expert blind to every judge's decision ($1{,}273$ rubric items, of which the expert decided $1{,}085$ and skipped $188$ as undecidable from the run folder or not applicable). Agreement is computed on the items both the expert and the judge decided; $n$ gives that count. Over-credit, judge \emph{pass} on an item the expert marked as failed; under-credit, judge \emph{fail} on an item the expert marked as passed. Abstention, fraction of all items the judge returned as undecidable. Intervals are $95\%$ cluster-bootstrap intervals over runs ($10^{4}$ resamples). The lower block splits the adopted judge by rubric subsection (general items) and by item severity.}\label{tab:judge}
{\small\setlength{\tabcolsep}{3.5pt}
\begin{tabular}{lN{1014}ccccc}
\toprule
Judge & \multicolumn{1}{c}{$n$} & Accuracy & Cohen's $\kappa$ & Over-credit & Under-credit & Abstention \\
\midrule
Claude Sonnet 5 (adopted) & 922 & 0.87 (0.85--0.90) & 0.67 (0.61--0.73) & 0.27 (0.20--0.34) & 0.07 & 0.20 \\
Claude Opus 5 & 1014 & 0.87 (0.85--0.89) & 0.66 (0.61--0.71) & 0.33 (0.27--0.38) & 0.05 & 0.10 \\
Gemini 3.1 Pro & 960 & 0.86 (0.85--0.88) & 0.67 (0.61--0.72) & 0.23 (0.17--0.27) & 0.10 & 0.17 \\
\midrule
\multicolumn{7}{l}{\emph{Adopted judge by rubric stratum}} \\
\quad Input understanding & 129 & 0.98 (0.95--1.00) & 0.87 (0.70--1.00) & 0.21 & 0.00 & 0.16 \\
\quad Statistical plotting & 67 & 0.88 (0.81--0.94) & 0.75 (0.57--0.88) & 0.09 & 0.14 & 0.16 \\
\quad Task-specific items (pooled) & 209 & 0.89 (0.83--0.93) & 0.72 (0.58--0.84) & 0.13 & 0.10 & 0.27 \\
\quad Quantification & 117 & 0.90 (0.84--0.95) & 0.69 (0.49--0.85) & 0.18 & 0.08 & 0.09 \\
\quad Reporting to the user & 73 & 0.84 (0.76--0.91) & 0.61 (0.40--0.79) & 0.12 & 0.27 & 0.25 \\
\quad Visualization & 230 & 0.83 (0.77--0.90) & 0.58 (0.44--0.71) & 0.37 & 0.08 & 0.12 \\
\quad Tool choice and use & 108 & 0.77 (0.69--0.84) & 0.47 (0.31--0.63) & 0.50 & 0.06 & 0.35 \\
\quad Critical items & 221 & 0.97 (0.95--0.99) & 0.52 (0.16--0.77) & 0.50 & 0.01 & 0.15 \\
\quad Major items & 365 & 0.89 (0.86--0.92) & 0.71 (0.63--0.79) & 0.28 & 0.04 & 0.24 \\
\quad Minor items & 336 & 0.79 (0.74--0.84) & 0.57 (0.47--0.67) & 0.26 & 0.17 & 0.19 \\
\bottomrule
\end{tabular}}
\end{table}

\begin{table}[H]\centering
\caption{\textbf{Rubric items that runs failed most often under expert review.} The twelve items with the highest fail rate under expert review among items decided in at least eight of the $20$ reviewed runs. \emph{Runs failed}, runs failed / runs decided by the expert; \emph{Judge agrees}, fraction of the runs decided by both on which the adopted judge returned the expert's verdict. Severity is the rubric's own weight class (critical $3$, major $2$, minor $1$).}\label{tab:faileditems}
{\small\setlength{\tabcolsep}{4pt}
\begin{tabular}{>{\raggedright\arraybackslash}p{7.7cm}lccc}
\toprule
Rubric item & Subsection & Severity & Runs failed & Judge agrees \\
\midrule
Did the agent provide a gallery/montage comparing different settings? & reporting & minor & 20/20 & 0.80 \\
Did the agent exclude saturated pixels from mean intensity calculations? & feature extraction & major & 8/8 & 1.00 \\
Did the agent check for clipped/saturated pixels before quantification? & input understanding & major & 16/17 & 0.92 \\
Did the agent provide a draft figure legend for each visualizations? & visualization & minor & 17/20 & 0.63 \\
Is the color map represented (color bar)? & visualization & minor & 15/18 & 0.73 \\
Did the agent suggest follow-up analyses based on findings? & quantification & minor & 16/20 & 0.79 \\
Did the agent suggest better alternatives for future runs? & reporting & minor & 15/20 & 0.89 \\
Is a scale bar present on the final visualization? & visualization & major & 15/20 & 1.00 \\
Did the agent use a quantitative metric to pick the best parameters? & tool use & major & 12/16 & 0.85 \\
Is the generated code documented well enough to serve as a tutorial? & reporting & minor & 14/19 & 0.87 \\
Did the agent disclose any non-linear adjustments during the figure making process (e.g., Gamma)? & visualization & major & 13/18 & 0.50 \\
Did the agent estimate parameters from visual features rather than using hard-coded defaults? & tool use & major & 13/19 & 0.25 \\
\bottomrule
\end{tabular}}
\end{table}

\FloatBarrier
\section{Token, runtime and cost accounting}\label{note:6}\label{app:usage}
Section~\ref{sec:cost} defines the efficiency measures and the two sources of cost. A monetary figure depends on
which vendor, tier or routing layer serves the model, and prices change. The
cost figures therefore state their source, and the underlying token counts are
always reported, from which any harness--model configuration can be re-priced. Provider billing is used for the Claude Code configurations and the DeepSeek-V4-Pro configuration because the harness spreads a run over multiple sessions, which metered client-side counts undercount and the billing ledger does not. Metered token usage is priced at the rates in Table~\ref{tab:prices}, with cached input
tokens priced at the cache-read rate and cache writes at the full input rate. This under-prices only Opus~5 on DeepSeek Harness, the one configuration whose
provider surcharges cache writes and whose cost is not taken from billing. Fresh
and cache-write tokens are $5\%$ of its input, which at Anthropic's
$1.25\times$ cache-write rate caps the unpriced surcharge at \$$0.25$ per run
on average and \$$0.94$ for the heaviest run.

Table~\ref{tab:telemetry} states, per agent, the fraction of runs exposing each signal. Input token counts include cached reads, which keeps totals comparable across agents whose providers bill cached reads differently.

Every harness--model configuration was run at the reasoning-effort tier
\emph{medium}, and the tier actually requested is written into each run's manifest. Four
harnesses expose a control for it directly: Claude Code through its
\texttt{--effort} flag, Codex through the \texttt{model\_reasoning\_effort}
configuration key, DeepSeek Harness through a route-level \texttt{reasoning} setting in its
provider patch, and Biomni through the OpenRouter \texttt{reasoning.effort}
field sent with every call. Agentic-J takes it from the
\texttt{IMAGENTJ\_REASONING\_EFFORT} environment variable, which a per-role
block in its mounted configuration can override; the tier each role actually
used is read back from its debug log. CopilotJ has no setting for it; our
driver injects the field through the pass-through arguments of its OpenAI
client and writes the applied value to the run log, which records whether the
route accepted the field. Table~\ref{tab:telemetry} lists, per harness, the signals exposed, where each
token count is read from and the medians.

\begin{table}[tbp]\centering
\caption{\textbf{Usage signals of the six agents on GPT-5.6 Sol under the brief instruction.} $n$, runs per agent; \emph{Tokens}, \emph{Tool calls}, \emph{Exec.}, fraction of runs for which the harness exposes a token count, a tool-call count and a count of executed code blocks; wall-clock runtime is measured by the wrapper for every run and is the one measure available for every agent. \emph{Source}, where the token count is read from. Medians are over runs that expose the signal; token medians are cache-inclusive input plus output. The reasoning-effort tier recorded in the run manifests is \emph{medium} for every harness (how it reaches each model is described in the text). A signal a harness does not expose is recorded as missing and excluded from the medians. DeepSeek Harness reports no tool-call count; Agentic-J (two runs) and CopilotJ (one run) exposed no counts for runs that the wall-clock limit killed.}\label{tab:telemetry}
{\small\setlength{\tabcolsep}{3.5pt}
\begin{tabular*}{\linewidth}{@{\extracolsep{\fill}}@{}lN{48}N{100\%}N{100\%}N{100\%}lN{2,171k}N{108}N{11}@{}}
\toprule
Harness & \multicolumn{1}{c}{$n$} & \multicolumn{1}{c}{Tokens} & \multicolumn{1}{c}{\shortstack{Tool\\calls}} & \multicolumn{1}{c}{Exec.} & \shortstack[l]{Token\\source} & \multicolumn{1}{c}{\shortstack{Med.\\tokens}} & \multicolumn{1}{c}{\shortstack{Med.\\calls}} & \multicolumn{1}{c}{\shortstack{Med.\\min}} \\
\midrule
Claude Code & 48 & 100\% & 100\% & 0\% & CLI result event & 982k & 39 & 11 \\
Codex & 48 & 100\% & 100\% & 0\% & CLI turn events (45); character estimate (3) & 2,171k & 15 & 10 \\
DeepSeek H. & 48 & 100\% & 0\% & 0\% & session usage events & 1,256k & -- & 10 \\
Biomni & 48 & 100\% & 100\% & 100\% & provider usage metadata & 293k & 10 & 7 \\
Agentic-J & 48 & 96\% & 96\% & 0\% & session totals & 3,029k & 108 & 32 \\
CopilotJ & 48 & 98\% & 98\% & 0\% & driver log & 271k & 22 & 7 \\
\bottomrule
\end{tabular*}}
\end{table}

\begin{table}[htbp]
\centering
\caption{\textbf{Models, API identifiers and list prices.} Identifiers are the
OpenRouter model slugs used in every run; the two DeepSeek models were served
as the 20260423 preview snapshots. Prices are the providers' list prices (USD
per million tokens) applied to metered token usage for every harness--model
configuration whose cost is not taken from provider billing, as verified at
the time of the study. The DeepSeek-V4-Pro rates are those of the provider
OpenRouter routed the model to; the V4-Pro and Claude Code costs reported in
the paper are taken from provider billing. Cached input is
billed at the cache-read rate; cache writes at the full input rate. The judge
model's usage is accounted and priced separately from the agents'.}\label{tab:prices}
{\small
\begin{tabular}{llN{0.00}N{0.000}N{00.00}}
\toprule
Model & API identifier & \multicolumn{1}{c}{Input} & \multicolumn{1}{c}{Cached input} & \multicolumn{1}{c}{Output} \\
\midrule
GPT-5.6~Sol             & \texttt{openai/gpt-5.6-sol}          & 2.00 & 0.200 & 10.00 \\
Claude Opus 5           & \texttt{anthropic/claude-opus-5}     & 5.00 & 0.500 & 25.00 \\
Claude Sonnet 5 (judge) & \texttt{anthropic/claude-sonnet-5}   & 2.00 & 0.200 & 10.00 \\
Kimi K2.6               & \texttt{moonshotai/kimi-k2.6}        & 0.95 & 0.160 & 4.00 \\
GLM-5.1                 & \texttt{z-ai/glm-5.1}                & 0.97 & 0.179 & 3.04 \\
DeepSeek-V4-Flash       & \texttt{deepseek/deepseek-v4-flash}  & 0.09 & 0.018 & 0.18 \\
DeepSeek-V4-Pro         & \texttt{deepseek/deepseek-v4-pro}    & 1.03 & 0.086 & 2.06 \\
\bottomrule
\end{tabular}}
\end{table}

\FloatBarrier
\section{Failure case studies}\label{note:7}\label{app:vignettes}
Three runs from the GPT-5.6~Sol study show how an analysis can look sound and still reach a wrong result. The judge rated all three procedurally sound (process scores $0.71$ to $0.87$), yet all three scored at or near zero on the outcome, and in each case the error is caught by comparing the delivered files with the reference, which a reading of the run would miss. The three cases are a report that contradicts the delivered table, a standard pipeline with an uncalibrated puncta definition, and a promise to finish that never produced the required file. Each case is taken from the archived trace, the executed code and the delivered files.

\paragraph{Case 1: mismatch between the delivered table and the report.}
CopilotJ on the 3D oncogenic puncta task (run \texttt{run\_20260829\_\allowbreak 062108};
process $0.71$, outcome $0.00$). The agent drove Fiji through $14$ macro calls
and $7$ Python executions, wrote a QC overlay for each of the $27$ stacks and
closed with ``Completed the full 3D analysis of all 27 two-channel confocal
stacks: 1,609 nuclei quantified, 27/27 images processed''. Its own final
validation step had printed the same figure of $1{,}609$. The per-nucleus table
it delivered holds $266$ nuclei, between $2$ and $23$ per stack (median $10$),
against $2{,}062$ in the reference (median $85$ per stack), and the
per-condition puncta counts it reports run in the wrong order (3DBDmut
$199$, wild type $94$, C129A/H144A $29$, against $23$, $36$ and $12$ in the
reference), inverting the biological conclusion. The report and the
file disagree because the validation counted a different intermediate than
the one finally written, and nothing compared the two. A nucleus count one eighth of the reference with a median nuclear volume of $775\,\mu$m$^3$ indicates that neighboring nuclei were merged, more than a stricter segmentation criterion could account for, and the $27$ overlays that would have shown it were written but never read.

\paragraph{Case 2: a sound pipeline with an uncalibrated puncta definition.}
Codex on the same task (run \texttt{run\_20260829\_\allowbreak 054044}; process $0.87$,
outcome $0.02$). The pipeline is a standard one, with Cellpose nuclei on the Hoechst channel,
a per-nucleus 3D mask, smoothed GFP, local maxima above a shell-estimated
background, and connected components accepted as puncta when they contain
at least $12$ voxels, occupy at most $14\,\mu$m$^3$ and stay within
$1.5\,\mu$m of the peak. It delivers a per-nucleus table, a per-punctum
table, condition summaries, overlays and a methods file.
The report gives $1{,}572$ nuclei, $34\%$, $21\%$ and $5\%$ puncta-positive
nuclei for wild type, 3DBDmut and C129A/H144A, and ``an approximately 91\%
reduction versus WT'' for the double mutant. The ordering is right (rank
metric $1.0$; per-image counts correlate with the reference at $r=0.85$) and
the magnitude is wrong by a factor of thirty, a mean of $1.2$ puncta per
wild-type nucleus against $36$ in the reference, whose distribution is heavy
tailed (a quarter of wild-type nuclei carry more than $30$). The detector
accepted only isolated droplets and discarded the dense clusters that
carry most of the reference count, and the thresholds that decide this
($15$ counts above background, $12$ voxels) were set once and never checked
against the images.

\paragraph{Case 3: a completion claim in the future tense.}
DeepSeek Harness on the NF-$\kappa$B translocation task (run
\texttt{run\_20260828\_\allowbreak 200015}; process $0.80$, outcome $0$, no required output file).
In $186$ seconds the agent wrote a complete $211$-line pipeline (plate-map
parsing, nuclear and cytoplasmic segmentation, per-well nucleus-to-cytoplasm
ratio, dose--response fit, a Kruskal--Wallis test, QC overlays and summary
tables) and a preview image, then ended its turn with: ``Analysis is still running
across all 96 wells. I'll continue automatically and produce the required
CSVs, QC overlays, dose-response plots, morphology analysis, and summary
report in the specified output directory.'' The harness exited normally at
that point; no process continued, and the per-well summary that the task
requires was never written. The judge, reading the script, rated the
methodology sound. An agent cannot work after its final turn, and a closing message that
promises continuation is indistinguishable from the outside from a claim of
completion. The run counts as a failure only because a missing required
output file scores zero.

\end{document}